\documentclass[pdflatex,sn-mathphys-num]{sn-jnl}

\usepackage{graphicx}%
\usepackage{multirow}%
\usepackage{amsmath,amssymb,amsfonts}%
\usepackage{amsthm}%
\usepackage{mathrsfs}%
\usepackage[title]{appendix}%
\usepackage[table]{xcolor}%
\usepackage{colortbl}
\definecolor{rowgray}{gray}{0.92}

\usepackage{pifont}
\usepackage{xcolor}
\newcommand{\xmark}{\textcolor{red}{\small\ding{55}}}

\usepackage{textcomp}%
\usepackage{manyfoot}%
\usepackage{booktabs}%
\usepackage{array}
\usepackage{algorithm}%
\usepackage{algorithmicx}%
\usepackage{algpseudocode}%
\usepackage{listings}%
\usepackage{subcaption}
\usepackage{underscore}

\usepackage{makecell} 

\theoremstyle{thmstyleone}%

\theoremstyle{thmstyletwo}%

\theoremstyle{thmstylethree}%
\newcommand{\imgplaceholder}[2]{
  \fbox{\parbox[c][0.12\textwidth][c]{#1}{\centering\footnotesize #2}}
}

\newcommand{\incimg}[2]{
  \IfFileExists{#2}{\includegraphics[width=#1]{#2}}{\imgplaceholder{#1}{#2}}
}

\begin{document}

\title[Article Title]{A Mixture of Experts Foundation Model for Scanning Electron Microscopy Image Analysis}

\author*[1]{Sk Miraj Ahmed}\email{sahmed3@bnl.gov}

\author[1]{Yuewei Lin}\email{\{ywlin}

\author[1]{Chuntian Cao}\email{ccao}

\author[1]{Shinjae Yoo}\email{sjyoo}

\author[2]{Xinpei Wu}\email{xwu4}
\author[2]{Won-Il Lee}\email{wlee2}
\author[2]{Nikhil Tiwale}\email{ntiwale\}@bnl.gov}

\author[3]{Dan N. Le}\email{\{dan.le}
\author[3]{Thi Thu Huong Chu}\email{thithuhuong.chu}
\author[3]{ Jiyoung Kim}\email{jiyoung.kim\}@utdallas.edu}

\author[2]{ Kevin G. Yager}\email{\{kyager}

\author[2]{Chang-Yong Nam}\email{cynam\}@bnl.gov}


\affil*[1]{\orgdiv{Computing and Data Sciences}, \orgname{Brookhaven National Laboratory}, \city{Upton}, \postcode{11973}, \state{New York}, \country{USA}}

\affil[2]{\orgdiv{Center for Functional Nanomaterials}, \orgname{Brookhaven National Laboratory}, \city{Upton}, \postcode{11973}, \state{New York}, \country{USA}}


\affil[3]{\orgdiv{Department of Material Sciences and Engineering}, \orgname{The University of Texas at Dallas}, \city{Richardson}, \postcode{75080}, \state{Texas}, \country{USA}}

\abstract{Scanning Electron Microscopy (SEM) is indispensable in modern materials science, enabling high-resolution imaging across a wide range of structural, chemical, and functional investigations. However, SEM imaging remains constrained by task-specific models and labor-intensive acquisition processes that limit its scalability across diverse applications. Here, we introduce the first foundation model for SEM images, pretrained on a large corpus of multi-instrument, multi-condition scientific micrographs, enabling generalization across diverse material systems and imaging conditions. Leveraging a self-supervised transformer architecture, our model learns rich and transferable representations that can be fine-tuned or adapted to a wide range of downstream tasks. As a compelling demonstration, we focus on defocus-to-focus image translation—an essential yet underexplored challenge in automated microscopy pipelines. Our method not only restores focused detail from defocused inputs without paired supervision but also outperforms state-of-the-art techniques across multiple evaluation metrics. This work lays the groundwork for a new class of adaptable SEM models, accelerating materials discovery by bridging foundational representation learning with real-world imaging needs.}

\keywords{Scanning electron microscopy, Foundation Model, Masked autoencoder, Mixture of Experts, Defocus-to-focus SEM restoration , metrology}

\maketitle

\section{Introduction}\label{sec:intro}

Scanning Electron Microscopy (SEM) is a cornerstone of ion- and electron-based materials characterization and a critical metrology \cite{postek2001critical,orji2018metrology} tool in advanced semiconductor manufacturing, particularly for extreme ultraviolet (EUV) lithography and resist pattern \cite{kumar2025resist, lorusso2022metrology} inspection. In modern process nodes, SEM is routinely used to quantify critical dimensions (CD), line-edge and line-width roughness (LER/LWR) \cite{lorusso2018unbiased, orji2021spectral}, stochastic defectivity, resist collapse, and pattern fidelity at sub-10 nm length scales—measurements that directly govern device performance, yield learning, and process control loops. However, these measurements are intrinsically coupled to imaging fidelity: even mild defocus, astigmatism, charging, beam-induced damage, or stage drift can introduce systematic bias in edge localization, roughness estimation, and power spectral density (PSD) analysis \cite{schubert2024deepfocus, maraghechi2019correction, abaidi2025analytical}. These effects are especially pronounced for EUV photoresists and other ion-sensitive materials, where low-dose imaging \cite{lorusso2022metrology, chung2025true, park2025deep} is mandatory and repeated acquisitions for focus bracketing or manual correction are often infeasible.

Despite the centrality of SEM in these workflows, SEM-based analysis remains dominated by narrowly tailored, task-specific algorithms and expert-driven acquisition protocols \cite{schubert2024deepfocus,MyScopeSEMArtefacts}. In practice, achieving usable images requires careful manual tuning of focus, stigmation, dwell time, and beam current—often guided by operator experience rather than objective criteria—and conservative imaging settings that sacrifice resolution to avoid charging or resist damage. As pattern dimensions shrink and stochastic variability increases, this reliance on ideal imaging conditions becomes a fundamental bottleneck: defocus and noise not only degrade visual interpretability, but also propagate non-trivially into downstream metrology, leading to inconsistent CD estimates, inflated roughness metrics, and reduced sensitivity to subtle process variations \cite{lorusso2018unbiased, abaidi2025analytical, orji2018metrology}. The diversity of SEM instruments, detectors, operating voltages, and sample types further complicates the development of robust, transferable analysis pipelines, limiting scalability across tools, fabs, and materials systems.

Collectively, these challenges highlight a growing mismatch between the scale and complexity of modern SEM data and the predominantly handcrafted nature of existing analysis approaches. As semiconductor manufacturing and ion-domain materials science increasingly demand high-throughput, automated, and statistically robust inspection, there is a pressing need for generalizable computational models that can tolerate realistic imaging imperfections, adapt across instruments and conditions, and decouple downstream analysis from strict acquisition constraints.
Recent advances in self-supervised learning \cite{jaiswal2020survey} and vision transformers (ViTs) \cite{dosovitskiy2020image} have fundamentally reshaped representation learning in natural images, enabling the emergence of large-scale foundation models \cite{bommasani2021opportunities} that learn transferable visual abstractions from unlabeled data \cite{he2022masked, caron2021emerging, chen2021empirical} and adapt to downstream tasks with minimal supervision. These models have demonstrated remarkable robustness to noise, resolution changes, and task variation—properties that are highly desirable for scientific imaging. Yet, despite the central role of Scanning Electron Microscopy (SEM) in materials science and semiconductor manufacturing, comparable foundation models  have not been developed for SEM data. This gap is nontrivial: SEM images differ substantially from natural images in terms of contrast mechanisms, noise statistics, texture distributions, and acquisition artifacts \cite{abaidi2025analytical, maraghechi2019correction}, while labeled data for SEM-specific tasks such as metrology or defect inspection remain scarce, expensive, and often instrument-dependent. As a result, the benefits of modern representation learning have yet to translate into practical, general-purpose tools for SEM analysis.

In this work, we introduce the first foundation model tailored specifically for SEM image data. Our approach is built on a masked autoencoding (MAE) framework, in which a ViT-Large backbone \cite{he2022masked,dosovitskiy2020image} is pretrained on 125,000 unlabeled SEM images spanning a broad range of materials systems, instruments, magnifications, and imaging conditions. To further enhance adaptability across the heterogeneous visual regimes encountered in SEM—ranging from smooth resist patterns to highly textured or noisy microstructures—we augment the transformer architecture with a Mixture of Experts (MoE) \cite{shazeer2017outrageously},  mechanism. By integrating sparse expert routing \cite{fedus2022switch} into the transformer blocks, the model can dynamically allocate capacity based on input-specific characteristics, effectively expanding the model to 1.7 billion parameters while maintaining efficient inference. This design allows different experts to specialize for variations in texture, resolution, and noise, addressing a key limitation of monolithic architectures in SEM settings.

As a concrete and practically relevant downstream application, we focus on defocus-to-focus image conversion, a critical capability for automating SEM workflows. In routine SEM operation—particularly for EUV resist inspection and ion-sensitive materials—defocus artifacts frequently arise due to sample drift, charging, astigmatism, or conservative low-dose acquisition settings. These artifacts not only degrade visual interpretability, but also introduce systematic bias into downstream metrology, affecting critical dimension (CD) estimation, roughness metrics, and frequency-domain analyses. Because acquiring paired real defocus–focus images is labor-intensive \cite{schubert2024deepfocus} and often infeasible, we fine-tune our pretrained foundation model using synthetically generated defocus–focus pairs produced by a physics-inspired forward image formation model. This simulator captures realistic blur anisotropy and signal-dependent noise without requiring per-image PSF estimation.

To enable fair, controlled, and reproducible evaluation, we adopt a fixed synthetic data generation protocol in which a randomly selected but fixed set of real SEM images is degraded using a seeded defocus and noise process. This allows us to benchmark performance under both synthetic-to-synthetic (S$\rightarrow$S) and synthetic-to-real (S$\rightarrow$R) settings, isolating the impact of SEM-aligned pretraining and architectural design on restoration quality. Importantly, this protocol avoids reliance on scarce real defocus–focus supervision or ad hoc calibration, while remaining faithful to the imaging imperfections encountered in practice.

Taken together, this work establishes a foundation-model paradigm \cite{bommasani2021opportunities, he2022masked} for SEM imaging, demonstrating that large-scale self-supervised learning—when carefully adapted to the physics, statistics, and operational constraints of electron microscopy—can yield general-purpose representations that support robust downstream analysis. By bridging modern representation learning with microscopy automation, our approach moves toward scalable, instrument-agnostic SEM workflows and contributes to the broader goal of accelerating materials characterization and discovery through intelligent imaging.

Our main contributions are summarized as follows:
\begin{itemize}

\item \textbf{MAE-Pretrained ViT with Block-Wise MoE Encoder.}
We propose a SEM foundation model trained via masked autoencoding that integrates a Mixture-of-Experts (MoE) mechanism at the transformer-block level. Sparse routing enables expert specialization for diverse SEM patterns while maintaining efficient inference.

\item \textbf{Physics-Grounded Synthetic Defocus and Noise Modeling.}
We adopt an elliptical Airy PSF model with Poisson--Gaussian noise and physically plausible parameter ranges to synthesize defocus–focus pairs, yielding a controllable and reproducible simulator for training and evaluation without requiring real-image PSF estimation.

\item \textbf{Controlled Synthetic-to-Synthetic and Synthetic-to-Real Protocols.}
We fine-tune and evaluate the model using fixed, seeded degradation processes on a fixed set of real SEM images, enabling consistent S$\rightarrow$S and S$\rightarrow$R benchmarking and isolating the benefits of SEM-aligned pretraining and MoE specialization.

\item \textbf{Improved Unsupervised Representation Quality via MAE + MoE.}
We show that augmenting a masked autoencoder with a Mixture-of-Experts encoder yields more discriminative SEM representations than MAE alone, as evidenced by consistently improved clustering performance on unlabeled SEM datasets.

\item \textbf{Measurement-Preserving Simulation-to-Real Transfer.}
We demonstrate that the learned SEM representations preserve critical metrology-relevant quantities, including CD and roughness statistics, on real SEM images—even when fine-tuning relies solely on synthetic defocus–focus supervision—highlighting robustness under limited or imperfect real data.

\end{itemize}

\begin{figure}[t]
    \centering
    \includegraphics[width=\linewidth]{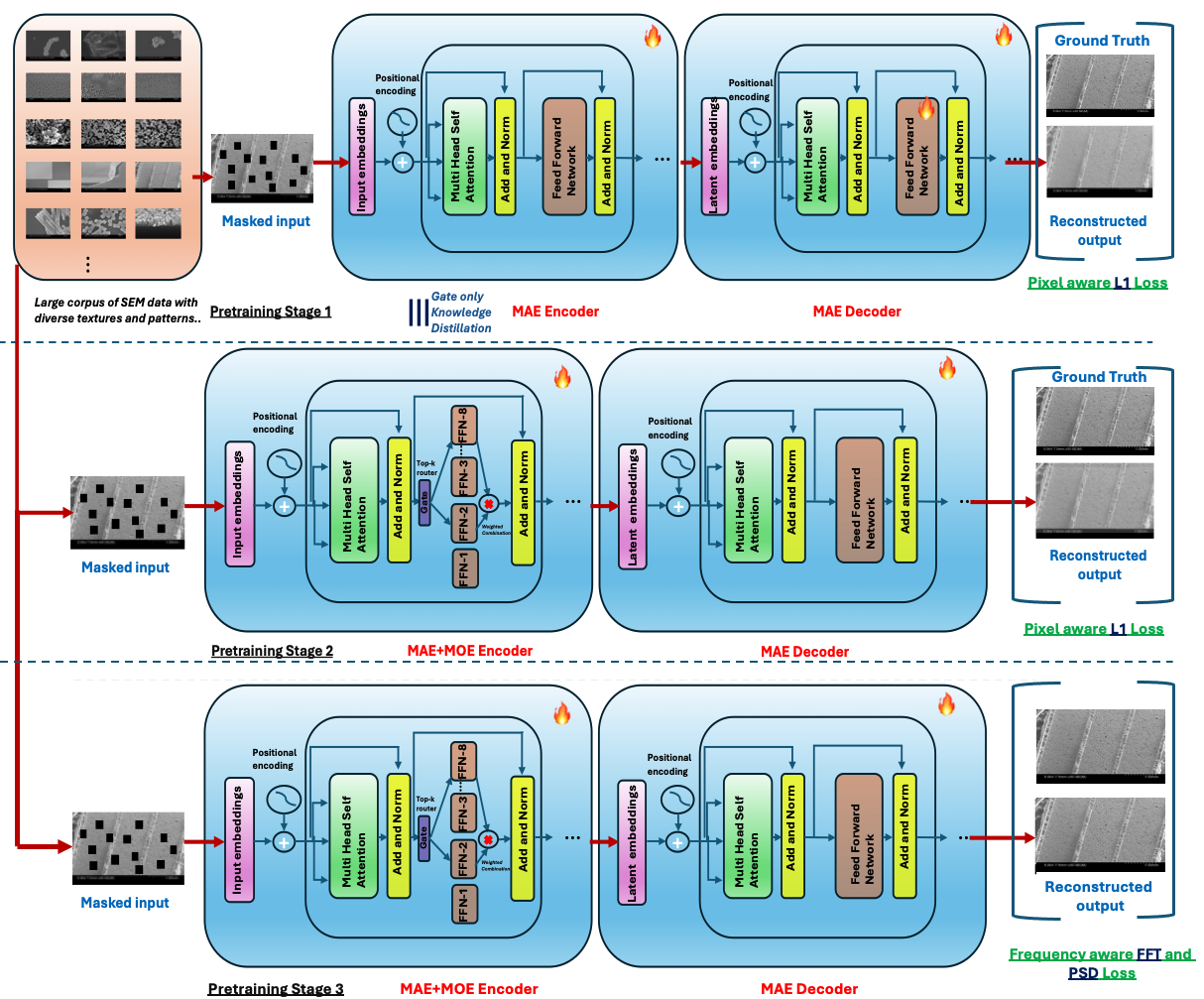}
    \caption{
\textbf{Overview of the proposed SEM foundation model pretraining framework.}
The framework consists of four stages:
(1) large-scale self-supervised pretraining of a ViT-MAE \emph{teacher} on diverse unlabeled SEM images;
(2) knowledge distillation into an MAE+MoE \emph{student} with multiple experts and top-$k$ gating to enable conditional computation and specialization;
(3) \textbf{frequency-aware masked reconstruction} to further adapt the foundation model to SEM-specific spectral statistics by emphasizing high-frequency/PSD-consistent detail recovery during masked prediction.
}
    \label{fig:sem_fm_framework}
\end{figure}

\begin{figure}[t]
    \centering
    \includegraphics[width=\linewidth]{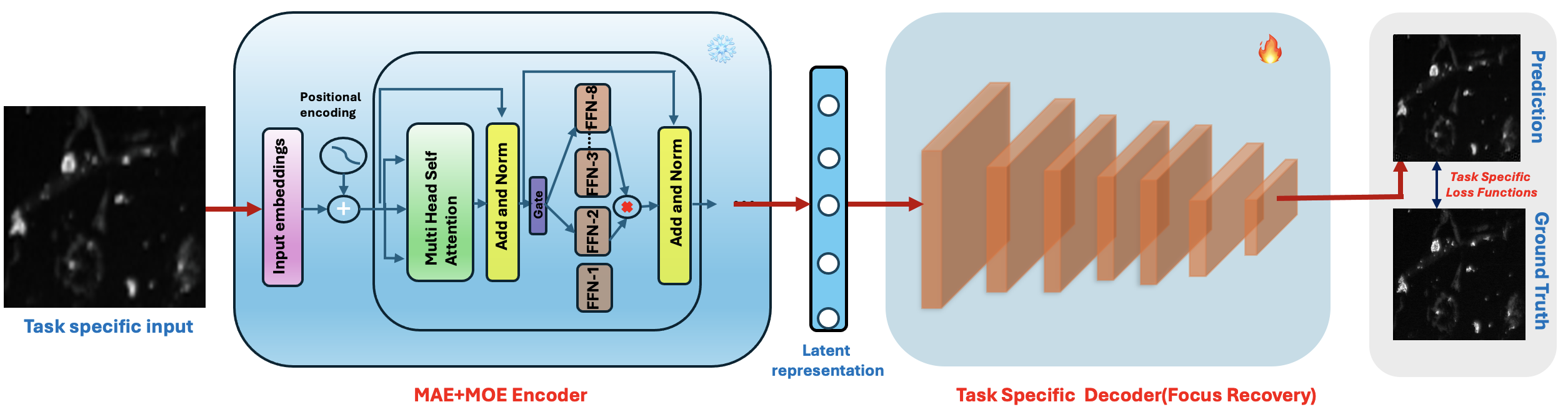}
    \caption{
\textbf{Downstream task adaptation.}
The pretrained SEM foundation encoder can be paired with \emph{any} task-specific decoder head (e.g., restoration, segmentation, or measurement prediction) depending on the application. In this work, we instantiate the downstream task as \textbf{SEM defocus-to-focus refocusing} and compare multiple decoder choices; empirically, reusing the \textbf{pretrained masked-reconstruction decoder} yields the best performance for refocusing, providing the most accurate and stable metrology on real SEM data. For adaptation, we compare the predicted focused image to the ground-truth focused reference under different loss configurations and fine-tune by updating \textbf{only the decoder} while keeping the encoder fixed.
}
    \label{fig:sem_fm_dt}
\end{figure}

\section{Experimental setup and Results}

\subsection{Datasets and Evaluation Protocol}
To evaluate our SEM foundation model and its downstream generalization capability, we curate a multi-instrument, multi-condition corpus of 125{,}000 unlabeled SEM images for self-supervised pretraining. The pretraining data spans diverse materials systems (e.g., metals, ceramics, polymers) and imaging conditions (e.g., varying beam energies, detector types, working distances, and sample topographies), capturing the heterogeneous distributions encountered in practical SEM workflows.

\paragraph{Pretraining details.}
We pretrain our SEM foundation model using a three-stage curriculum. All stages use the same unlabeled SEM image corpus and train/validation splits; images are converted to RGB, bottom-cropped to remove acquisition artifacts, resized to $224\times224$, and augmented with random resized cropping, full-range rotation, horizontal/vertical flips, and mild color jitter. The masking ratio is fixed to $0.75$ throughout. In Stage-1, we pretrain a ViT-MAE-Large model (patch size $16$) using the standard masked reconstruction objective \cite{he2022masked} for $400$ epochs with batch size $64$ on $2$  H100 GPUs, optimized with AdamW \cite{loshchilov2017decoupled} (learning rate $1\times10^{-4}$, no weight decay). In Stage-2, we replace the feed-forward networks in each transformer block with an MoE module with $8$ experts and top-1 routing, initialize experts by copying pretrained FFN weights, and distill from the Stage-1 MAE while training only the gating parameters using the same reconstruction loss plus a load-balancing regularizer (weight $0.01$). In Stage-3, all parameters are unfrozen and training continues with additional frequency-aware structural losses (each weighted $0.1$) and the same load-balancing regularizer, using AdamW with learning rate $1\times10^{-5}$ and weight decay $0.05$ for $400$ epochs under distributed training on $2$ GPUs. Model selection is based on minimum validation loss, and PSNR/SSIM are logged for monitoring only.

\subsection{Baseline Methods}
\label{sec:baselines}

We compare our approach against a diverse set of baselines spanning classical SEM restoration, non-learning denoising methods, self-supervised low-shot models, and task-specific deep learning approaches. All methods are evaluated under the same synthetic-to-real (S$\rightarrow$R) or synthetic-to-synthetic (S$\rightarrow$S) protocols described earlier, using identical preprocessing and evaluation metrics whenever applicable.

\paragraph{Classical deconvolution baselines.}
We include two widely used physics-based deconvolution methods for SEM image restoration: Richardson--Lucy (RL) \cite{richardson1972bayesian, lucy1974iterative}and Wiener deconvolution \cite{wiener1949extrapolation}. Both methods operate in a zero-shot manner and do not involve learning or data-driven adaptation.

\textbf{Richardson--Lucy deconvolution (RL).\cite{lucy1974iterative}}
RL is an iterative maximum-likelihood method derived under a Poisson noise model. It is applied using a fixed point spread function (PSF) instantiated as an elliptical Airy kernel, consistent with the optics-based forward model used for synthetic defocus generation. The PSF is kept fixed across all test images, reflecting realistic deployment scenarios in which the true PSF is unknown and per-image estimation is unavailable. No test-time supervision or parameter tuning is performed.

\textbf{Wiener deconvolution.\cite{wiener1949extrapolation}}
Wiener deconvolution is a frequency-domain approach that balances blur inversion and noise suppression under a stationary Gaussian noise assumption. We use the same fixed Airy-based PSF as for RL to ensure consistency across classical baselines. As with RL, Wiener deconvolution is applied deterministically without test-time tuning.

For both deconvolution methods, the PSF is normalized to unit energy and shared across all images. This avoids unrealistically favorable per-image PSF estimation and provides a conservative yet deployment-faithful baseline.

\paragraph{Classical denoising baseline.}
\textbf{BM3D. \cite{dabov2007image}}
We include BM3D as a strong non-learning denoising baseline to assess the extent to which noise suppression alone can improve SEM image quality without explicitly modeling defocus blur. BM3D is applied to grayscale images in a zero-shot manner using a test-set--agnostic noise estimation strategy. As BM3D does not account for the defocus PSF, it serves as an informative reference for the limitations of denoising-only restoration in recovering high-frequency structure lost to blur.

\paragraph{Self-supervised denoising baselines.}

\textbf{Noise2Noise (N2N). \cite{lehtinen2018noise2noise}}
We include a Noise2Noise-style self-supervised denoising baseline that learns from pairs of independently corrupted observations of the same underlying signal. The model is trained using standard patch-based supervision and applied in a fully feed-forward manner at test time.

\textbf{Noise2Void (N2V). \cite{krull2019noise2void}}
We also include a Noise2Void-style blind-spot denoising baseline, which learns from single noisy observations by predicting masked pixels from their spatial context. The model is trained using a standard blind-spot corruption scheme and applied feed-forward at inference.

Both N2N and N2V operate purely as denoisers and do not explicitly model defocus blur, providing a complementary comparison to deconvolution-based and refocusing-specific methods.

\paragraph{Task-specific deep learning baseline.}
\textbf{MRN (Multi-Resolution Refocus Network). \cite{na2021deep}}
We compare against MRN, a task-specific deep refocusing model based on a multi-scale encoder--decoder architecture. MRN predicts refocused outputs at multiple spatial resolutions using pyramid-based supervision and represents prior deep learning approaches designed specifically for SEM defocus-to-focus restoration.

\paragraph{Foundation model baselines.}
\textbf{ViT-MAE (ImageNet-pretrained). \cite{he2022masked}}
We include a masked autoencoder with a Vision Transformer (ViT) encoder pretrained on ImageNet \cite{russakovsky2015imagenet} as a domain-mismatched foundation baseline. This model reflects the common practice of transferring large-scale natural image representations to scientific imaging tasks. The pretrained model is adapted to SEM refocusing using the same training protocol and losses as our method, isolating the effect of pretraining domain mismatch.

\noindent\textbf{ViT-MAE + MoE (zero-shot(ours)).}
We evaluate our SEM-pretrained ViT-MAE with a Mixture-of-Experts (MoE) decoder in a zero-shot setting, where the model is applied directly to defocused SEM images without any task-specific fine-tuning or access to defocus--focus pairs. This baseline probes the intrinsic representational capability of the SEM-aligned foundation model and its expert routing mechanism under purely unsupervised deployment.

\noindent\textbf{ViT-MAE + MoE (ours).}
Our final model uses the same SEM-pretrained ViT-MAE with an MoE decoder but allows low-shot adaptation using a small number of defocus--focus image pairs. During adaptation, the decoder experts and routing mechanism are optimized jointly, enabling the model to specialize to the target defocus characteristics. All competing baselines are evaluated under the same low-shot supervision regime, highlighting the advantage of controlled adaptation on top of a domain-aligned foundation model.

\paragraph{Evaluation protocol and metrics.}
Restored images are evaluated using a combination of full-reference image-quality metrics, no-reference perceptual metrics, and measurement-oriented metrology metrics. When ground-truth focused images are available, we report peak signal-to-noise ratio (PSNR) \cite{sara2019image} , which measures pixel-wise reconstruction fidelity, and structural similarity index (SSIM) \cite{wang2004image}, which assesses local luminance, contrast, and structural consistency. To better capture perceptual and structural agreement beyond pixel-wise similarity, we additionally report LPIPS \cite{zhang2018unreasonable}, which compares deep feature representations between restored and reference images and is more sensitive to edge alignment and texture preservation.

For scenarios where a reliable reference image is unavailable or imperfectly aligned, we report NIQE \cite{mittal2012making} as a no-reference quality indicator. NIQE measures deviation from natural-image statistics and serves as a diagnostic for severe artifacts or unnatural distortions; however, it is not optimized for SEM imagery and is therefore interpreted as a secondary indicator rather than a primary objective.

To directly assess suitability for downstream microscopy analysis, we compute a suite of metrology metrics on all restored images using the same measurement pipeline applied to ground-truth focused SEM images. These include critical dimension (CD) error \cite{azarnouche2012unbiased}, which quantifies absolute bias in measured feature width; CD standard deviation, which reflects measurement stability; line width roughness (LWR) and line edge roughness (LER), which capture spatial fluctuations along feature boundaries; and power spectral density (PSD)–based diagnostics, which evaluate frequency-domain consistency of edge roughness. These metrics are particularly sensitive to subtle edge distortions that may be visually inconspicuous but are critical for semiconductor manufacturing and materials characterization.

This comprehensive evaluation protocol enables us to contrast physics-based inversion methods, denoising-only approaches, task-specific deep models, and foundation-model-based methods under identical experimental conditions, while explicitly distinguishing visual fidelity from measurement reliability.

\noindent\textbf{Remark 1.}
ViT-MAE models operate on fixed-size inputs (224$\times$224 pixels in our implementation), which imposes a practical constraint when processing high-resolution SEM images. While larger input resolutions are in principle possible, they incur prohibitive computational and memory costs due to the quadratic scaling of self-attention, making full-image inference infeasible under realistic resource budgets.
To enable evaluation on full-resolution SEM images, we therefore adopt a sliding-window inference strategy. The input image is decomposed into overlapping 224$\times$224 patches, which are processed independently by the model and subsequently stitched together using a Hann window with an 8-pixel overlap. This overlap-and-weighted-averaging scheme mitigates boundary artifacts and ensures smooth spatial transitions between adjacent patches without introducing post-hoc sharpening or heuristic blending.
Although patch-wise processing may discard some long-range contextual information, our results indicate that the learned representation is sufficiently robust to preserve both local structure and global consistency at evaluation time. Exploring full-scale SEM training and inference—via more memory-efficient attention mechanisms, hierarchical tokenization, or multi-resolution processing—remains an important direction for future work to further reduce information loss and improve global coherence.

\noindent\textbf{Remark 2.}
BM3D, Noise2Noise (N2N), and Noise2Void (N2V) are primarily denoising-oriented baselines and do not explicitly model or invert defocus blur. Nevertheless, they are widely used in SEM practice as first-line enhancement tools. Their inclusion allows us to disentangle improvements arising from noise suppression alone from those due to genuine defocus inversion and structural recovery.

\subsection{Baseline Experimental Settings and Hyperparameters}
\label{sec:baseline_settings}

To ensure a fair and deployment-faithful comparison, we evaluate all baselines under a unified preprocessing and pairing protocol, and we avoid any test-time tuning that would provide oracle access to the held-out images. In particular, all classical and learning-based baselines operate on grayscale SEM images and use the same bottom-crop operation (crop fraction $0.0667$) to remove scan-dependent text information prior to restoration and evaluation. When paired focused references are available, defocused and focused images are paired by sorted index order (rather than filename matching), reflecting the practical setting where metadata may be inconsistent. We report PSNR/SSIM/LPIPS when references exist and additionally report NIQE as a no-reference diagnostic on all restored outputs.


\paragraph{Synthetic training/evaluation data and defocus model with parameter ranges.}
To generate realistic synthetic defocus for training and evaluation, we adopt a physics-inspired forward model based on an \emph{elliptical Airy disk point spread function (PSF)} augmented with intensity scaling and SEM-like noise processes. Defocus and astigmatism are modeled using an anisotropic Airy PSF parameterized by horizontal and vertical radii $(R_x, R_y)$, a rotation angle $\theta$, and a sharpness exponent $\beta$. The PSF is constructed by rotating spatial coordinates, computing an elliptical radial distance, evaluating the squared first-order Bessel function response, and normalizing to unit energy; convolution uses reflective boundary conditions to avoid edge artifacts. To account for realistic SEM intensity variations and acquisition noise, the blurred image is further transformed via multiplicative gain $a$ and additive bias $b$, followed by dose-aware Poisson noise (electron counting statistics) and additive Gaussian noise with standard deviation $\sigma$ (sensor/electronic noise).

Because well-aligned real defocus/focus pairs are scarce, we synthesize defocus--focus training data by applying the above degradations on-the-fly to clean focused SEM images, and we use the same synthetic pipeline to train learning-based baselines (e.g., Noise2Noise and Noise2Void) for fair comparison. Rather than estimating PSF parameters from real data or performing per-image calibration, we sample forward-model parameters independently from broad, physically plausible ranges (defined in the Method section) to cover diverse SEM operating conditions:
\[
R_x \sim \mathcal{U}(1,30),\quad
R_y \sim \mathcal{U}(1,30),\quad
\beta \sim \mathcal{U}(1.9,2.0),\quad
\theta \sim \mathcal{U}(0,3.14),
\]
\[
a \sim \mathcal{U}(0.99,1.1),\quad
b \sim \mathcal{U}(1,25),\quad
\sigma \sim \mathcal{U}(1,10),\quad
\text{dose} \sim \mathcal{U}(1,50),
\]
where $R_x,R_y$ control PSF radii (pixels), $\beta$ controls Airy tail sharpness, $\theta$ is the PSF orientation, $a,b$ model intensity scaling/offset, $\sigma$ is the Gaussian noise level, and \text{dose} controls Poisson shot-noise strength (per Eqn.~\ref{fr-model}). This wide-coverage sampling avoids over-specialization to a single blur configuration and promotes robust simulation-to-real transfer without relying on real-image supervision or PSF fitting.

For training, we select a small set of \textbf{10 real focused SEM images} and synthesize many defocused observations by sampling parameters per patch from the above ranges. For evaluation, we independently select \textbf{100 real focused SEM images} and generate corresponding synthetic defocused inputs using the same forward model, ensuring strict separation between training and test images. During controlled synthetic evaluation, we also report a representative setting obtained by fixing each parameter to the midpoint of its range. Overall, training across a diverse family of defocus and noise realizations enables the model (and learning-based baselines) to learn invariances relevant to SEM image formation while maintaining a realistic small-data adaptation regime.

\paragraph{Real SEM evaluation data.}
We also evaluate on \textbf{real} defocus--focus SEM pairs acquired from lithographically patterned resist structures spanning multiple material systems and feature scales. The dataset includes chemically amplified resists (CAR) patterned by EUV lithography, electron-beam lithography (EBL) patterns in ZEP resist, and Zn-containing hybrid resist patterns fabricated via molecular layer deposition and EUV exposure. Collectively, these samples cover a broad range of critical dimensions and pattern densities representative of advanced semiconductor manufacturing. Images are acquired under controlled SEM settings at multiple magnifications and focus offsets, including in-focus, under-focus, and over-focus conditions, yielding realistic defocused inputs paired with focused references. This diversity of materials, patterning processes, and focus conditions provides a challenging benchmark for assessing simulation-to-real transfer and real-world defocus-to-focus restoration performance. Detailed fabrication protocols and imaging conditions are provided in the Supplementary Material.

\paragraph{Classical baselines (fixed, deployment-style parameters).}
For classical restoration methods we use fixed hyperparameters applied uniformly across all images, mirroring how these operators are used in practice and preventing test-set tuning. Richardson--Lucy (RL) deconvolution is run for 30 iterations. Wiener deconvolution uses a fixed balance parameter of 0.01. Both RL and Wiener use a single elliptical Airy PSF with $R_x=R_y=15.5$ pixels and $\beta=1.95$ (midpoint values of the synthetic bounds), with a fixed unrotated kernel ($\theta=0$) for a conservative isotropic baseline. The PSF kernel size is set to $k=\lceil 6\cdot \max(R_x,R_y)\rceil$ and enforced to be odd.

\paragraph{BM3D (noise-level selection).}
BM3D requires a noise standard deviation parameter $\sigma$. To avoid oracle tuning while accommodating varying acquisition noise, we estimate $\sigma$ per image using a robust MAD-based estimator on a high-pass residual (difference from a $3{\times}3$ median-filtered image), operating in the normalized $[0,1]$ intensity scale. All other BM3D settings use standard library defaults, ensuring a deterministic and reproducible denoising-only baseline.

\paragraph{Self-supervised denoisers.}
Noise2Noise (N2N) trains a U-Net on randomly cropped patches using an MSE (or L1) loss between two independent noisy realizations of the same underlying clean patch. Noise2Void (N2V) trains a U-Net using blind-spot masking with a masking ratio of 0.1 and computes loss only on masked pixels. Both are trained on synthetically degraded patches generated from clean SEM images using the parameter ranges above and then applied feed-forward to real defocused SEM images at test time.

\paragraph{Task-specific deep baseline.}
MRN (MultiScaleRefocusNet) is trained using a loss $\ell_1$ across coarse/intermediate/fine resolutions and optimized with Adam using the learning rate $5\times 10^{-5}$, batch size 4, and 500 epochs. The best checkpoint is selected with the lowest validation loss and evaluated under the same preprocessing and metric pipeline.

Overall, these settings reflect realistic SEM usage: classical methods are applied with fixed operator-level parameters, while learning-based baselines rely on physics-guided synthetic degradations rather than extensive real paired supervision, consistent with the practical scarcity of well-aligned defocus/focus SEM pairs.

\subsection{Training Setup for Our Method (downstream task)}

\paragraph{Synthetic defocus and noise model.}
Our method employs the same physics-inspired synthetic defocus and noise model as described for the learning-based baselines. Specifically, elliptical Airy PSF parameters and signal-dependent noise are sampled using the \emph{same parameter choices and distributions as defined in the baseline experimental setup}, ensuring full consistency between baselines and our approach. No additional tuning of PSF or noise parameters is performed for our model.

\paragraph{Fine-tuning protocol.}
We fine-tune our SEM-pretrained ViT-MAE (and its MoE variant) using a low-shot adaptation setting. Only a small set of in-focus SEM images is used, from which synthetic defocused counterparts are generated on-the-fly. No real defocus/focus pairs are used for training unless explicitly stated.

\paragraph{Optimization and training details.}
During fine-tuning, the ViT encoder is frozen and only the decoder parameters are updated. Models are trained for 400 epochs using the AdamW optimizer with a learning rate of $1\times10^{-4}$ and zero weight decay. We use a batch size of 1 per GPU and train with 2 H100 GPUs using distributed data parallelism.

\paragraph{Input preprocessing.}
All images are converted to grayscale, normalized using the ViT-MAE image processor, and cropped by removing the bottom $6.7\%$ of each image to avoid SEM-specific artifacts. Random multi-scale crops of sizes $\{128,256,512,1024\}$ are used during training, while validation uses a fixed center crop.

\paragraph{Mixture-of-Experts configuration.}
For the ViT-MAE+MoE variant, we replace each encoder feed-forward network with an 8-expert Mixture-of-Experts module using top-1 routing. Expert weights are initialized from the pretrained MAE decoder, and a lightweight entropy-based load-balancing regularizer is applied during training.


\paragraph{Loss functions.}
We evaluate two loss configurations for our ViT-MAE+MoE model. 
In the first variant, denoted as \emph{ViT-MAE+MoE ($\ell_1$)} in the tables, training uses only the Charbonnier (robust $\ell_1$) reconstruction loss in Eq.~\ref{eq:charb} between the restored and ground-truth focused images. 
In the second variant, denoted as \emph{ViT-MAE+MoE ($\ell_1$+TV)}, we optimize the full objective in Eq.~\ref{eq:total_loss} by augmenting Eq.~\ref{eq:charb} with an edge-consistency loss computed from image gradients (Eq.~\ref{eq:edge_loss}, weighted by $\lambda_e{=}3$) and a total variation (TV) regularizer (Eq.~\ref{eq:tv}, weighted by $\lambda_{tv}{=}10$), which suppresses spurious high-frequency artifacts while preserving sharp structural boundaries. 
We choose $\lambda_e$ and $\lambda_{tv}$ to balance the relative magnitudes of the three terms so that no single component dominates optimization. 
All other training settings are kept identical between the two variants.

\subsection{Foundation Model Analysis: MAE vs.\ MAE+MoE}
\label{sec:foundation-model-analysis}

We first analyze the benefits of integrating a Mixture-of-Experts (MoE) decoder into the pretrained MAE foundation model. Despite the increase in capacity (from 400M to 1.7B parameters), the inference time of MAE+MoE remains comparable to the baseline MAE due to top-1 gating that activates only a single expert per token. As summarized in Table~\ref{tab:mae_vs_moe_clustering}, we assess unlabeled clustering quality with three intrinsic diagnostics: \emph{Silhouette} (cosine; $\uparrow$) \cite{rousseeuw1987silhouettes} capturing the trade-off between within-cluster cohesion and nearest-cluster separation; \emph{Davies--Bouldin} (DBI; $\downarrow$) \cite{davies2009cluster} measuring the worst-case ratio of intra-cluster scatter to inter-centroid distance; and \emph{Calinski--Harabasz} (CH; $\uparrow$) \cite{calinski1974dendrite} quantifying between- versus within-cluster dispersion. On the 31{,}628-image benchmark with $K{=}10$, MAE+MoE improves Silhouette from 0.0643 to 0.1014 (+57.7\%), reduces DBI from 2.9303 to 2.5721 ($-12.2\%$), and increases CH from 1462.40 to 2027.61 (+38.7\%), indicating tighter, better-separated clusters while maintaining comparable inference latency via top-1 routing.


\begin{table}[t]
\centering
\caption{Clustering performance comparing ViT-MAE Large vs.\ ViT-MAE + MoE (8 experts). Higher is better for Silhouette and Calinski--Harabasz; lower is better for Davies--Bouldin.}
\label{tab:mae_vs_moe_clustering}
\renewcommand{\arraystretch}{1.2}
\small 

\begin{tabular*}{\columnwidth}{@{\extracolsep{\fill}} l ccc @{}}
\toprule
\textbf{Model} 
& \makecell{\textbf{Silhouette}\\\textbf{(cosine)} $\uparrow$} 
& \makecell{\textbf{Davies--}\\\textbf{Bouldin} $\downarrow$} 
& \makecell{\textbf{Calinski--}\\\textbf{Harabasz} $\uparrow$} \\
\midrule
ViT-MAE Large & 0.0643 & 2.9303 & 1462.40 \\
ViT-MAE Large + MoE (8E) & \textbf{0.1014} & \textbf{2.5721} & \textbf{2027.61} \\
\midrule
\textit{Relative change vs.\ MAE} & {+57.7\%} & {$-12.2\%$} & {+38.7\%} \\
\bottomrule
\end{tabular*}
\end{table}

\subsection{Synthetic-to-Synthetic Evaluation Analysis}
\label{sec:s2s_analysis}
Table \ref{tab:s2s_visual} reports synthetic-to-synthetic (S$\rightarrow$S) restoration results under a fully controlled optics-simulated defocus setting, where both training and evaluation are performed on synthetically blurred SEM images. This protocol isolates the behavior of each method under known degradation, allowing a focused comparison of visual fidelity, structural consistency, and frequency preservation without confounding real-world acquisition variability.
Classical deconvolution methods exhibit limited robustness in this setting. While Richardson–Lucy improves SSIM relative to the defocused input, it substantially degrades PSNR and LPIPS, indicating over-amplification of noise and ringing artifacts. Wiener filtering performs poorly across all metrics, particularly in NIQE, reflecting strong sensitivity to noise-model mismatch. These results highlight the limitations of fixed, non-adaptive optics inversion when applied to realistic SEM-like degradations.
Learning-based denoising methods such as BM3D, Noise2Void, and Noise2Noise provide moderate improvements over the defocused input, particularly in SSIM and NIQE. However, these approaches primarily suppress noise rather than explicitly modeling defocus blur, resulting in limited gains in high-frequency recovery as reflected by LPIPS and PSNR. The task-specific MRN refocusing network improves PSNR relative to denoising baselines but does not consistently outperform them across perceptual metrics, suggesting sensitivity to the specific blur configuration used during training.
The ImageNet-pretrained ViT-MAE performs poorly in this controlled setting, achieving the lowest PSNR and SSIM among all methods. This underscores the domain mismatch between natural image pretraining and SEM image statistics, even when evaluated under synthetic degradations.
In contrast, our SEM-pretrained ViT-MAE with a Mixture-of-Experts (MoE) encoder demonstrates strong performance even without downstream adaptation. In the zero-shot setting, the model achieves competitive PSNR and SSIM while substantially outperforming all baselines in NIQE, indicating superior perceptual realism and frequency consistency. Allowing limited adaptation further improves performance across all metrics, yielding the best PSNR, SSIM, and LPIPS scores in the table while maintaining low NIQE. Notably, the consistent improvement in frequency- and perception-sensitive metrics suggests that the MAE+MoE architecture captures structural priors aligned with SEM image formation, rather than merely fitting the synthetic degradation.
Overall, these results show that combining masked autoencoding with expert specialization leads to representations that are both visually faithful and structurally robust under controlled defocus. This establishes a strong foundation for subsequent synthetic-to-real transfer, where maintaining measurement-critical structures is essential despite incomplete or imperfect supervision.

\begin{table}[t]
\centering
\caption{Synthetic-to-synthetic (S$\rightarrow$S) evaluation using optics-simulated defocus.Both training and evaluation are performed on synthetically blurred SEM images. Metrics reflect visual fidelity and frequency preservation under fully controlled conditions.}
\label{tab:s2s_visual}
\renewcommand{\arraystretch}{1.2}
\small 

\begin{tabular*}{\columnwidth}{@{\extracolsep{\fill}} lcccc @{}}
\toprule
\textbf{Method} 
& \textbf{PSNR $\uparrow$} 
& \textbf{SSIM $\uparrow$} 
& \textbf{LPIPS $\downarrow$} 
& \textbf{NIQE $\downarrow$} \\
\midrule
Defocused Input                & 18.3 & 0.31 & 0.57  & 11.8 \\
Richardson--Lucy               & 15.9 & 0.51 & 0.76 & 10.9 \\
Wiener Filter                  & 14.5 & 0.30 & 0.83 & 19.7 \\
BM3D                           & 18.7 & 0.44 & 0.51 & 14.4 \\
Noise2Void (N2V)               & 18.9 & 0.55 & 0.56 & 9.4 \\
Noise2Noise (N2N)              & 18.8 & 0.55 & 0.60 & 9.6 \\
MRN       & 19.9 & 0.49 & 0.62 & 10.5 \\
ViT-MAE (ImageNet)              & 12.5   & 0.17   &0.50   & 13.7   \\
\addlinespace 
\rowcolor{rowgray}
\makecell[l]{ViT-MAE + MoE \\ (ours zero shot)} & 18.3 & 0.50 & 0.52 & 5.3 \\
\rowcolor{rowgray}
\makecell[l]{ \textbf{ViT-MAE + MoE} \\ \textbf{(ours few shot)}}   & \textbf{20.2} & \textbf{0.57} & \textbf{0.50} & \textbf{5.3} \\
\bottomrule
\end{tabular*}
\end{table}

\subsection{Synthetic-to-Real Evaluation with a Single Real Reference}
Table \ref{tab:real_one_pair} reports defocus-to-focus restoration results on real fast-scan SEM images using a single slow-scan image as the in-focus reference (data acquisition process details in supplementary). All learning-based models are trained exclusively on ten synthetically generated defocus–focus pairs and evaluated directly on real defocused inputs, forming a challenging synthetic-to-real (S$\rightarrow$R) setting.
While some methods achieve lower NIQE by producing visually smoother outputs, this often comes at the expense of structural fidelity. In this context, reference-based perceptual metrics are more informative than no-reference natural-image quality measures. Accordingly, LPIPS is the most relevant indicator, as it directly measures structural agreement with the focused SEM reference.
Under this criterion, the proposed ViT-MAE + MoE model achieves the lowest LPIPS while simultaneously improving PSNR and SSIM, indicating superior preservation of morphology and fine structures. The slightly higher NIQE reflects deviation from natural-image statistics rather than degradation of SEM-relevant content. These results demonstrate that the structural prior learned by the SEM foundation model enables reliable simulation-to-real transfer, even when trained solely on synthetic data.

\begin{table}[t]
\centering
\caption{Image-quality evaluation for defocus-to-focus SEM restoration using a single real slow-scan image as the in-focus reference.
All models are trained exclusively on 10 synthetically generated defocus–focus image pairs and evaluated on real fast-scan defocused SEM images.
Higher is better for PSNR and SSIM, while lower is better for LPIPS and NIQE.}
\label{tab:real_one_pair}
\begin{tabular*}{\columnwidth}{@{\extracolsep{\fill}} lcccc @{}}
\toprule
\textbf{Method} 
& \textbf{PSNR $\uparrow$} 
& \textbf{SSIM $\uparrow$} 
& \textbf{LPIPS $\downarrow$} 
& \textbf{NIQE $\downarrow$} \\
\midrule
Fast-scan Input (Defocused) 
& 17.7 & 0.62 & 0.35 & 7.9 \\

Richardson--Lucy 
& 15.8 & 0.53 & 0.37 & 11.2 \\

Wiener Filter 
& 11.9 & 0.23 & 0.54 & 11.3 \\

BM3D 
& 17.8 & 0.63 & 0.34 & 8.2 \\

Noise2Void (N2V) 
& 17.4 & 0.47 & 0.34 & 10.0 \\

Noise2Noise (N2N) 
& 17.2 & 0.46 & 0.34 & 11.1 \\

MRN
& \textbf{22.8} & 0.70 & 0.35 & 9.9 \\

\rowcolor{rowgray}
ViT-MAE (SEM-pre.) 
& 17.7 & 0.64 & 0.34  & \textbf{5.3} \\

\rowcolor{rowgray}
ViT-MAE + MoE (ours) 
& 22.6 & \textbf{0.70} & \textbf{0.32} & 9.9 \\
\bottomrule
\end{tabular*}
\end{table}

\begin{table}[t]
\centering
\caption{Metrology accuracy for defocus-to-focus SEM restoration under a synthetic-to-real (S$\rightarrow$R) protocol.Models are trained on synthetic defocus and evaluated on real defocused SEM images. Values are reported for two real images (Img1 / Img2). \textbf{CD(MAE)} is averaged over Img1 and Img2, and \textbf{Avg(MAE)} is computed by averaging the absolute error over all 10 entries (5 metrics $\times$ 2 images) with respect to the focused reference. Entries marked with \xmark \quad indicate cases where the metrology software (SMILE) failed to reliably detect edges due to poor image quality.
}
\label{tab:metrology}
\setlength{\tabcolsep}{1.2pt} 
\renewcommand{\arraystretch}{1.2}
\scriptsize

\begin{tabular*}{\columnwidth}{@{\extracolsep{\fill}} l cc cc cc cc cc c c @{}}
\toprule
\textbf{Method}
& \multicolumn{2}{c}{\textbf{CD}}
& \multicolumn{2}{c}{\makecell{\textbf{CD}\\\textbf{Std}}} 
& \multicolumn{2}{c}{\textbf{LWR}}
& \multicolumn{2}{c}{\textbf{LER}}
& \multicolumn{2}{c}{\textbf{PSD}}
& \makecell{\textbf{CD}\\\textbf{(MAE)}} 
& \makecell{\textbf{Avg}\\\textbf{(MAE)}} \\ 
\cmidrule(lr){2-3}\cmidrule(lr){4-5}\cmidrule(lr){6-7}
\cmidrule(lr){8-9}\cmidrule(lr){10-11}
& I1 & I2 & I1 & I2 & I1 & I2 & I1 & I2 & I1 & I2 & & \\
\midrule
Focused Input
& 16.3 & 9.7 & 0.4 & 0.6 & 4.2 & 2.2 & 3.3 & 1.6 & 4.2 & 2.2 & 0.00 & 0.00 \\
\midrule
Defocused Input
& 16.9 & 10.4 & 1.4 & 0.7 & 6.4 & 2.5 & 4.6 & 1.9 & 6.5 & 2.5 & 0.65 & 0.91 \\
Richardson--Lucy
& 17.4 & 10.7 & 1.4 & 0.8 & 4.3 & 1.4 & 3.0 & 1.1 & 4.4 & 1.4 & 1.05 & 0.60 \\
Wiener Filter            
& 18.0 & \xmark & 0.4 & \xmark & 9.3 & \xmark & 7.2 & \xmark & 9.4 & \xmark & \xmark & \xmark \\
BM3D
& 17.1 & 10.2 & 1.6 & 0.7 & 4.8 & 2.3 & 3.4 & 1.7 & 5.1 & 2.3 & 0.65 & 0.50 \\
Noise2Noise (N2N)        
& 17.8 & 10.6 & 1.6 & 0.8 & 4.4 & 1.4 & 3.1 & 1.1 & 4.7 & 1.4 & 1.3 & 0.70 \\
Noise2Void (N2V)
& 17.8 & 10.8 & 1.6 & 0.8 & 4.4 & 1.4 & 3.0 & 1.1 & 4.6 & 1.4 & 1.30 & 0.70 \\
MRN
& 19.7 & 11.7 & 1.4 & 0.6 & 4.0 & 1.4 & 2.9 & 1.1 & 4.1 & 1.4 & 2.70 & 0.92 \\
ViT-MAE (ImgNet)       
& 13.6 & \xmark & 1.1 & \xmark & 6.0 & \xmark & 4.9 & \xmark & 6.5 & \xmark & \xmark & \xmark \\
\rowcolor{rowgray}
ViT-MAE (SEM)
& 17.0 & 10.1 & 1.6 & 0.6 & 5.7 & 2.0 & 4.0 & 1.5 & 5.8 & 2.0 & 0.55 & 0.66 \\
\rowcolor{rowgray}
\textbf{ViT-MAE + MoE ($\ell_1$)}
& \textbf{16.1} & \textbf{9.7}
& \textbf{1.5} & \textbf{0.8}
& \textbf{5.4} & \textbf{2.0}
& \textbf{3.9} & \textbf{1.6}
& \textbf{5.7} & \textbf{2.0}
& \textbf{0.10} & \textbf{0.52} \\
\rowcolor{rowgray}
\textbf{ViT-MAE + MoE ($\ell_1$+TV)}
& \textbf{16.7} & \textbf{9.4}
& \textbf{1.7} & \textbf{0.8}
& \textbf{4.1} & \textbf{1.3}
& \textbf{3.0} & \textbf{1.0}
& \textbf{4.5} & \textbf{1.3}
& \textbf{0.35} & \textbf{0.53} \\
\bottomrule
\end{tabular*}
\end{table}

\begin{figure}[t]
\centering
\setlength{\tabcolsep}{0pt} 
\renewcommand{\arraystretch}{1.2}
\scriptsize

\begin{tabular*}{\columnwidth}{@{\extracolsep{\fill}} cccc @{}}
\toprule
\textbf{GT (Focused)} & \textbf{Defocus Input} & \textbf{Rich.--Lucy} & \textbf{Wiener} \\
\midrule
\includegraphics[width=0.23\columnwidth]{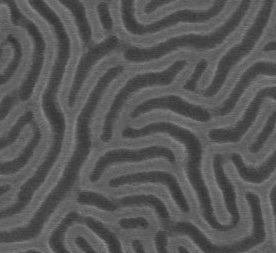} &
\includegraphics[width=0.23\columnwidth]{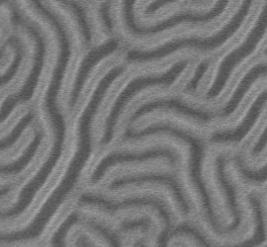} &
\includegraphics[width=0.23\columnwidth]{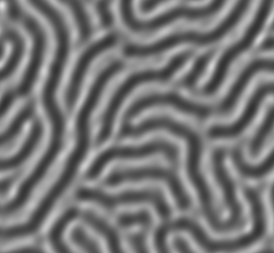} &
\includegraphics[width=0.23\columnwidth]{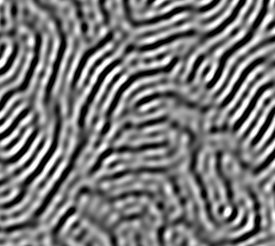} \\
\midrule
\textbf{BM3D} & \makecell{\textbf{Noise2Void}\\\textbf{(N2V)}} & \makecell{\textbf{Noise2Noise}\\\textbf{(N2N)}} & \textbf{MRN} \\
\midrule
\includegraphics[width=0.23\columnwidth]{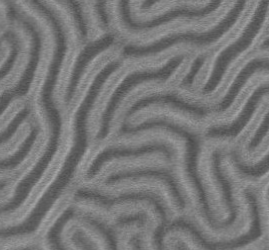} &
\includegraphics[width=0.23\columnwidth]{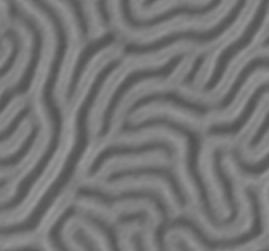} &
\includegraphics[width=0.23\columnwidth]{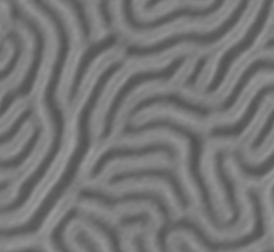} &
\includegraphics[width=0.23\columnwidth]{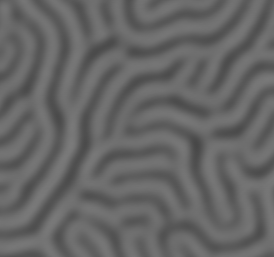} \\
\midrule
 \makecell{\textbf{ViT-MAE}\\\textbf{(ImageNet)}} & \makecell{\textbf{ViT-MAE}\\\textbf{(SEM)}} & \makecell{\textbf{ViT-MAE+MoE}\\\textbf{($\ell_1$)}} & \makecell{\textbf{ViT-MAE+MoE}\\\textbf{($\ell_1$+TV)}} \\
\midrule
\includegraphics[width=0.23\columnwidth]{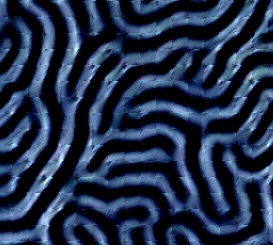} &
\includegraphics[width=0.23\columnwidth]{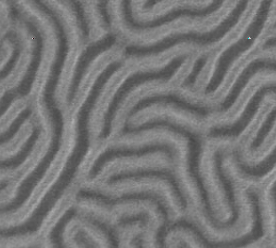} &
\includegraphics[width=0.23\columnwidth]{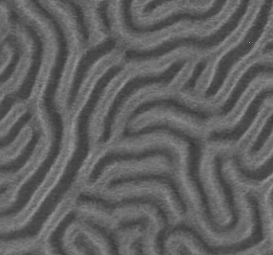} &
\includegraphics[width=0.23\columnwidth]{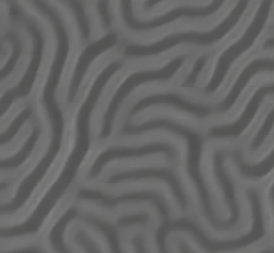} \\
\bottomrule
\end{tabular*}

\vspace{3pt}
\caption{Qualitative comparison for defocus-to-focus SEM restoration under the S$\rightarrow$R protocol. The focused slow-scan image is used as reference (GT), and the fast-scan defocused image is the input.
All remaining panels are restored outputs from classical baselines, denoising baselines, and learning-based methods.}
\label{fig:qual_grid_3x4}
\end{figure}

\paragraph{Qualitative comparison.}
Figure~\ref{fig:qual_grid_3x4} presents a qualitative comparison of defocus-to-focus restoration under the S$\rightarrow$R protocol. Classical deconvolution and denoising baselines either fail to fully recover high-frequency structure or introduce visible artifacts, while the ImageNet-pretrained ViT-MAE suffers from clear domain mismatch. The SEM-pretrained ViT-MAE improves visual realism but still exhibits residual blur and contrast inconsistencies. Among all methods, \textbf{ViT-MAE + MoE ($\ell_1$)} produces restorations that are visually closest to the focused ground-truth image, preserving edge sharpness, line geometry, and local texture with minimal artifacts. This qualitative observation is consistent with the quantitative results across all tables, where the $\ell_1$ MoE variant achieves the best overall balance between visual fidelity, perceptual similarity, and metrology accuracy.

\subsection{Metrology Accuracy under Synthetic-to-Real Transfer}
\label{sec:metrology_analysis}

Table~\ref{tab:metrology} is the \emph{most important quantitative result} in this work, as it directly evaluates whether defocus-to-focus restoration preserves \emph{measurement-critical quantities} on real SEM data. Unlike perceptual or image-quality metrics, the reported metrology measures---critical dimension (CD), CD standard deviation, line width roughness (LWR), line edge roughness (LER), and power spectral density (PSD)---determine whether restored images remain valid for downstream semiconductor process control and materials characterization.

All metrology measurements are computed using \textbf{SMILE} (SEM-Measured Image Lines Estimator), an open-source SEM image metrology toolkit widely used in EUV resist screening for extracting CD and roughness statistics from line/space and contact patterns~\cite{mochi2020open, mochi2021contacts}. We use the built-in \textbf{polynomial fitting} option provided in the SMILE-based metrology pipeline (as commonly done in resist-screening workflows) to fit measurement trends (e.g., dose-dependent CD/LWR curves) and obtain stable summary estimates under noisy SEM observations~\cite{develioglu2023euv}.

The evaluation is intentionally stringent. All models are trained exclusively on synthetically generated defocus--focus pairs, yet are evaluated on real defocused SEM images under a synthetic-to-real (S$\rightarrow$R) protocol. Errors are reported separately for two real images (I1/I2), and summarized via \textbf{CD(MAE)}, which averages CD error across both images, and \textbf{Avg(MAE)}, which averages absolute error across all ten entries (five metrics $\times$ two images). This setting exposes even subtle structural distortions that may be visually inconspicuous but lead to biased or unstable measurements.

Classical deconvolution and denoising baselines show limited reliability in this regime. While Richardson--Lucy and Wiener filtering can sharpen edges, they frequently introduce ringing or noise amplification that increases CD error or yields unstable measurements, with Wiener filtering failing to produce valid outputs for some cases. Denoising-based methods (BM3D, Noise2Noise, Noise2Void) suppress noise but do not explicitly model defocus blur, resulting in only modest improvements and persistent errors in CD, roughness, and PSD. The task-specific MRN improves several roughness-related metrics but exhibits the largest CD(MAE), indicating that visually sharp reconstructions can still induce systematic CD bias on real data. The ImageNet-pretrained ViT-MAE generalizes poorly, underscoring the inadequacy of natural-image priors for SEM metrology.

In contrast, the SEM-pretrained ViT-MAE substantially reduces metrology error across all metrics, demonstrating that self-supervised pretraining on large unlabeled SEM datasets learns structural priors aligned with SEM edge statistics and measurement pipelines. Building on this foundation, the proposed ViT-MAE + MoE models achieve the lowest summary errors, with consistent improvements in CD(MAE) and Avg(MAE) across both images. Notably, these gains are achieved despite training solely on synthetic data, highlighting strong simulation-to-real transfer.

Overall, Table~\ref{tab:metrology} shows that accurate SEM refocusing cannot be judged by visual fidelity alone. Preserving quantitative measurements requires SEM-specific representations, and the combination of masked autoencoding with expert specialization provides the most reliable preservation of CD and roughness statistics under real defocus conditions.

\begin{figure}[t]
    \centering
    \includegraphics[width=\linewidth]{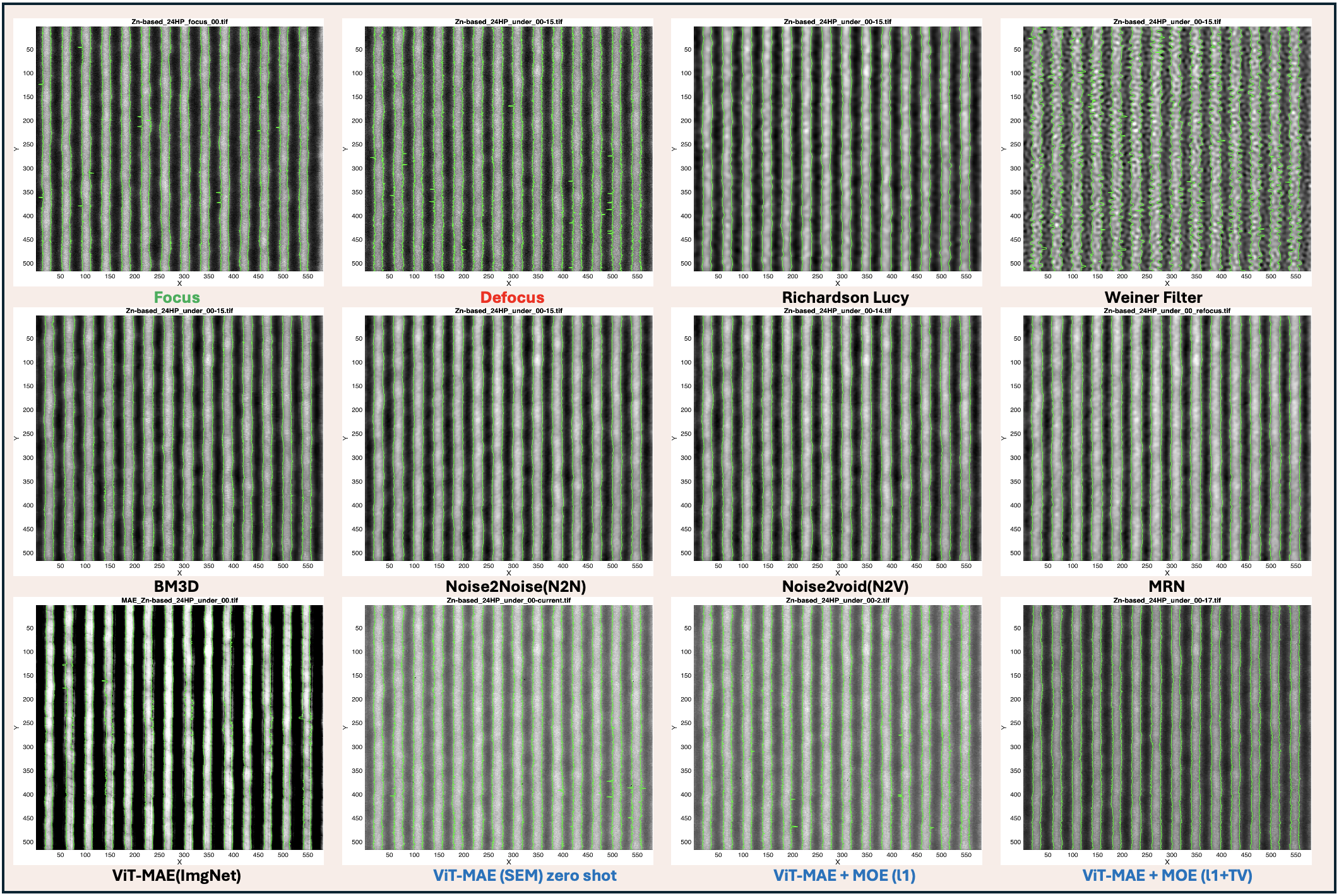}
    \caption{\textbf{SMILE-based line detection for metrology.} Visualization of the line/edge detections produced by the SMILE pipeline for the focused reference (top-left), the defocused input, representative baselines, and our restored outputs (bottom row, last three columns). While several methods yield visually sharp reconstructions, they can still induce mis-localized or inconsistent SMILE line fits that lead to biased CD/roughness measurements. In contrast, our outputs produce line detections that are most consistent with the focused reference, aligning with the improved metrology accuracy reported in Table~\ref{tab:metrology}.}
    \label{fig:vis_sem_lines}
\end{figure}

Moreover, as a qualitative complement to Table~\ref{tab:metrology}, Fig.~\ref{fig:vis_sem_lines} visualizes the \textbf{SMILE} line-detection outputs used for CD/roughness extraction. The overlays show that the proposed methods (bottom row, last three columns) yield line fits that are visually more stable and spatially consistent with the focused reference in the top-left, whereas several baselines produce irregular or locally shifted detections. Importantly, some competing restorations can appear visually sharp yet still lead to mis-localized edges and biased metrology, highlighting the disconnect between perceptual quality and measurement fidelity. In contrast, our reconstructions maintain both strong visual agreement with the focused reference and the closest correspondence to the reference metrology.

\begin{figure}[t]
    \centering
    \includegraphics[width=\linewidth]{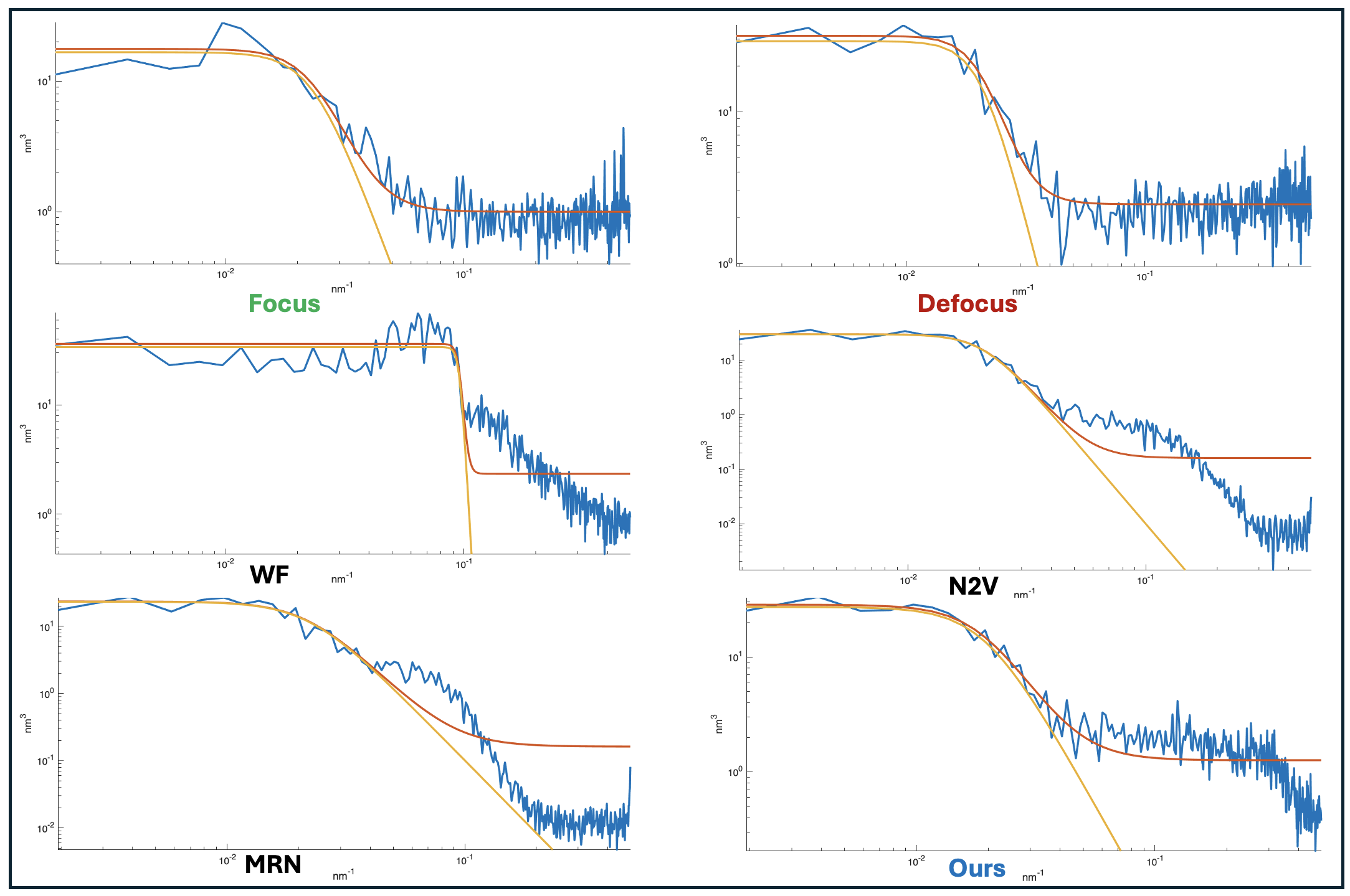}
    \caption{\textbf{LWR power spectral density (PSD) from SMILE.} LWR PSD curves directly exported from the SMILE metrology pipeline for the focused reference (F), defocused input (DF), Wiener filtering (WF), MRN, Noise2Void (N2V), and our method (subset shown for readability). Our reconstruction produces a PSD that is most coherent with the focused reference across the relevant frequency band, while other methods show larger spectral distortions indicative of biased roughness statistics. Residual high-frequency attenuation in our outputs can be adjusted by tuning the relative weights of the loss components during fine-tuning.}
    \label{fig:vis_sem_psd}
\end{figure}

We also directly export the \textbf{LWR power spectral density (PSD)} curves from the SMILE software and report them in Fig.~\ref{fig:vis_sem_psd}. For clarity, we visualize only the focused reference (F), the defocused input (DF), Wiener filtering (WF), MRN, Noise2Void (N2V), and our method. As shown in the PSD plots, our reconstruction yields a spectrum that is the most coherent with the focused reference across the relevant frequency range, whereas competing methods exhibit stronger deviations that indicate biased roughness statistics. We observe mild attenuation at very high frequencies in our results; in practice, this behavior can be further controlled by tuning the relative weights of the loss components (e.g., edge and TV terms) to trade off noise suppression and high-frequency detail preservation.

\section{Discussion}

To the best of our knowledge, this work presents the \emph{first large-scale foundation model trained specifically on scanning electron microscopy (SEM) data}. Unlike prior SEM restoration or enhancement methods that rely on task-specific architectures or limited supervised training, our approach leverages large-scale self-supervised pretraining to learn a general-purpose representation of SEM imagery, which can then be efficiently adapted to downstream tasks such as defocus-to-focus restoration.

Our results demonstrate that foundation-model-based adaptation is particularly effective under realistic experimental constraints, where paired real focused data are scarce or unavailable. Across synthetic-to-real (S$\rightarrow$R) and real-only evaluation protocols, the proposed method consistently outperforms classical physics-based deconvolution, denoising-oriented baselines, and task-specific deep networks. Importantly, these gains extend beyond standard perceptual metrics and translate into substantial improvements in metrology-relevant measurements, including critical dimension (CD) error, line-edge roughness (LER), and frequency-domain diagnostics.

A key insight from our study is that improvements in visual fidelity do not necessarily correspond to improvements in measurement accuracy. Several baselines achieve competitive PSNR or SSIM but introduce subtle structural distortions that degrade metrology outcomes. By contrast, the proposed foundation model preserves both spatial structure and spectral characteristics of SEM images, enabling reliable downstream measurement using the same pipelines applied to ground-truth focused data.

The realism of the training and evaluation setup plays a crucial role in this performance. Classical deconvolution methods require careful parameter tuning and are sensitive to mismatches between assumed and actual imaging conditions, while denoising-only methods are fundamentally limited in their ability to invert defocus blur. In contrast, our approach combines physics-inspired degradation modeling with representation learning, allowing the model to generalize across a wide range of defocus, noise, and acquisition settings with minimal real-data adaptation.

Despite these advantages, several limitations remain. The current framework operates on single images and does not exploit temporal or multi-frame information that may be available in certain SEM acquisition workflows. Additionally, while the degradation model captures a broad class of defocus and noise effects, further extensions could improve robustness to extreme imaging regimes or instrument-specific artifacts. Addressing these directions is a promising avenue for future work.

\section{Problem Formulation and Foundation Model Pretraining (Fig.~\ref{fig:sem_fm_framework})}
\label{sec:pretrain}

\paragraph{Setting and goal.}
Let $\mathcal{X}=\{x_i\}_{i=1}^N$ be a large corpus of \emph{unlabeled} SEM micrographs. For clarity, we describe the grayscale case
$x_i \in \mathbb{R}^{H\times W}$; extension to multi-channel SEM modalities (e.g., multi-detector or multi-energy acquisitions) is
straightforward by treating $x$ as a tensor in $\mathbb{R}^{C\times H\times W}$.
Our objective is to learn a general-purpose \emph{encoder} $f_\theta$ that maps an SEM image to a compact representation
\begin{equation}
z_i = f_\theta(x_i)\in\mathbb{R}^{D},
\end{equation}
such that $z_i$ is transferable: it can be adapted with limited labeled data and modest compute to diverse downstream SEM tasks
(e.g., denoising, defocus-to-focus restoration, segmentation, metrology-relevant regression/classification, anomaly detection).

\paragraph{Why self-supervised pretraining for SEM?}
In many SEM workflows, acquiring task-specific labels (e.g., pixel-level segmentations or metrology ground truth) is expensive and
instrument-dependent, while unlabeled images are abundant.
Self-supervised learning leverages this abundance by training models to solve \emph{pretext} objectives that do not require labels,
yet force the network to capture morphology, texture, edges, and other structure that is predictive for downstream tasks.
We adopt a three-stage strategy designed around three requirements that are particularly important in SEM:
(i) faithful recovery of fine spatial details (Stage~1),
(ii) specialization to heterogeneous image statistics (Stage~2), and
(iii) consistency in frequency-domain cues that correlate with SEM metrology (Stage~3).

\paragraph{Transformer-based encoder at a high level.}
Our encoder is a \emph{Vision Transformer} (ViT). Unlike convolutional networks that process an image through local filters,
a ViT first decomposes the image into \emph{patches} (small square regions), converts each patch into a vector (a ``token''),
and then repeatedly mixes information across all tokens using \emph{self-attention}.
Self-attention can be viewed as a learned, data-adaptive mechanism for deciding which regions of the image should
communicate with which other regions (e.g., long-range interactions between repeated patterns, edges spanning large distances,
or structures distributed over the field of view). This global interaction is valuable for SEM, where relevant structure is often
non-local (e.g., periodic lines/spaces, repeated textures, multi-scale defects).

\subsection{Stage 1: Masked Autoencoder (MAE) Pretraining}
\label{sec:stage1_mae}

\paragraph{Patchification and tokens.}
Given an image $x\in\mathbb{R}^{H\times W}$, we partition it into $P$ non-overlapping patches of size $s\times s$:
\begin{equation}
\{p_j\}_{j=1}^{P}, \qquad P=\frac{HW}{s^2}.
\end{equation}
Each patch $p_j$ is flattened and linearly projected into a $d$-dimensional token embedding; we denote the resulting token sequence
by $\{t_j\}_{j=1}^{P}$, with $t_j\in\mathbb{R}^{d}$.
Positional encodings are added so that the model retains knowledge of \emph{where} each patch came from.

\paragraph{Masking protocol.}
We randomly select a subset of patch indices $\mathcal{P}_m \subset \{1,\dots,P\}$ with masking ratio $\gamma\in(0,1)$.
The encoder observes only \emph{visible} tokens $\{t_j\}_{j\notin\mathcal{P}_m}$.
For pixel-level bookkeeping, let $M\in\{0,1\}^{H\times W}$ be the induced mask over pixels:
$M(u,v)=1$ if pixel $(u,v)$ lies inside a masked patch and $0$ otherwise.
(Equivalently, one may define a patch-level mask; both are consistent since patches are non-overlapping.)

\paragraph{Encoder--decoder reconstruction.}
The MAE consists of an encoder $f_\theta$ and a lightweight decoder $g_\phi$.
The encoder produces latent features from visible patches; the decoder receives these features along with \emph{mask tokens}
standing in for the missing patches, and reconstructs the full image:
\begin{equation}
\hat{x} = g_\phi\!\left( f_\theta(\{t_j\}_{j \notin \mathcal{P}_m}) \right) \in \mathbb{R}^{H\times W}.
\end{equation}

\paragraph{Masked reconstruction objective.}
Crucially, MAE training evaluates reconstruction error \emph{only where information was removed} (the masked region),
preventing the model from wasting capacity on copying visible pixels:
\begin{equation}
\mathcal{L}_{\text{MAE}}(x,\hat{x};M)
=
\frac{1}{\|M\|_1}
\sum_{u=1}^{H}\sum_{v=1}^{W} M(u,v)\, \big|\hat{x}(u,v)-x(u,v)\big|.
\label{eq:mae_loss}
\end{equation}
We use a normalized $\ell_1$ loss for robustness to outliers and the heavy-tailed intensity variations often observed in SEM.

\paragraph{Instantiation.}
The encoder $f_\theta$ is a ViT-Large backbone (24 transformer blocks, hidden size 1024) pretrained on $N=125{,}000$ unlabeled SEM images.
After Stage~1, $f_\theta$ serves as a strong generic SEM feature extractor, while the decoder is primarily a training scaffold.

\subsection{Stage 2: Mixture-of-Experts (MoE) Adaptation via Knowledge Distillation}
\label{sec:stage2_moe}

\paragraph{Motivation: heterogeneity in SEM data.}
SEM images vary widely across instruments, materials, magnifications, scan settings, and noise regimes, leading to heterogeneous
statistics (e.g., smooth regions vs.\ highly textured nanostructures; sharp edges vs.\ blurred defocus; low-dose noise vs.\ clean scans).
A single set of feed-forward layers may underfit this diversity.
Mixture-of-Experts (MoE) increases model capacity by maintaining multiple specialized sub-networks (``experts'') and learning to route
each token to the most appropriate expert(s).

\paragraph{MoE in a transformer block.}
In a standard transformer block, the \emph{feed-forward network} (FFN) transforms each token independently after attention mixing.
We replace the decoder FFN with an MoE layer:
\begin{equation}
\text{MoE}(h) \;=\; \sum_{k=1}^{K} \alpha_k(h)\, \text{FFN}_k(h),
\label{eq:moe}
\end{equation}
where $h\in\mathbb{R}^{d}$ is a token embedding, $\text{FFN}_k$ is the $k$-th expert, and $\alpha_k(h)\ge 0$ are routing weights
satisfying $\sum_{k=1}^{K}\alpha_k(h)=1$.
The gating network $G$ maps tokens to routing weights, e.g.,
\begin{equation}
\alpha(h) = \text{softmax}(G(h)) \in \mathbb{R}^{K}.
\end{equation}
(Top-$k$ routing is a common variant that selects only the largest $k$ weights for efficiency; our formulation covers both dense and top-$k$ routing.)

\paragraph{Knowledge distillation for stable MoE initialization.}
Directly training a higher-capacity MoE decoder from scratch can be unstable.
We therefore distill from a \emph{frozen} Stage~1 MAE teacher.
Let $\hat{x}^{\,T}$ be the teacher reconstruction and $\hat{x}^{\,S}$ be the MoE student reconstruction under the same mask $M$.
We apply distillation only over masked pixels:
\begin{equation}
\mathcal{L}_{\text{KD}}
=
\frac{1}{\|M\|_1}
\sum_{u,v} M(u,v)\, \big|\hat{x}^{\,S}(u,v)-\hat{x}^{\,T}(u,v)\big|.
\label{eq:kd_loss}
\end{equation}
Intuitively, the teacher provides a strong ``target'' for how missing SEM content should be inferred, while the student learns
how to distribute this inference across experts.

\paragraph{Expert load balancing.}
A common failure mode of MoE is \emph{expert collapse} \cite{shazeer2017outrageously}, where the gate routes most tokens to a small subset of experts, leaving others unused.
To encourage balanced utilization, we regularize the batch-averaged routing weights \cite{lepikhin2020gshard, fedus2022switch}.
Let $\alpha_{i,t,k}$ denote the routing weight to expert $k$ for token $t$ in sample $i$, for a batch of size $B$ and $P$ tokens:
\begin{equation}
\bar{\alpha}_k =
\frac{1}{BP}
\sum_{i=1}^{B}\sum_{t=1}^{P} \alpha_{i,t,k}.
\end{equation}
We penalize deviation from uniform usage using the standard quadratic form:
\begin{equation}
\mathcal{L}_{\text{LB}}
=
K \sum_{k=1}^{K} \bar{\alpha}_k^2,
\label{eq:lb_loss}
\end{equation}
whose minimum is achieved at $\bar{\alpha}_k = 1/K$ for all $k$.

\paragraph{Stage~2 objective.}
The Stage~2 training objective combines masked reconstruction, distillation, and load balancing:
\begin{equation}
\mathcal{L}_{\text{joint}}
=
\mathcal{L}_{\text{MAE}}
+
\lambda\,\mathcal{L}_{\text{KD}}
+
\mu\,\mathcal{L}_{\text{LB}},
\label{eq:stage2_obj}
\end{equation}
where $\lambda$ and $\mu$ control the strength of distillation and load balancing.
After Stage~2, the model has increased capacity and can adaptively specialize to different SEM regimes through expert routing.

\subsection{Stage 3: Frequency-Aware Masked Refinement}
\label{sec:stage3_freq}

\paragraph{Why frequency-domain refinement in SEM?}
Many SEM analyses depend not only on pixel-wise similarity but also on frequency-sensitive attributes:
line/space patterns induce strong spectral peaks; roughness and stochastic texture affect the power spectral density (PSD);
and blur or defocus suppresses high-frequency content.
Pixel-domain losses can yield visually plausible reconstructions that nonetheless distort these spectral cues, impacting metrology.
Stage~3 therefore adds frequency-aware penalties that explicitly constrain reconstruction quality in the Fourier domain.

\paragraph{Masked error in Fourier space.}
Let $\mathcal{F}(\cdot)$ denote the 2D discrete Fourier transform (DFT).
We define the masked reconstruction error and its spectrum:
\begin{equation}
e_M = M \odot (\hat{x} - x),
\qquad
E_M = \mathcal{F}(e_M),
\label{eq:masked_error_fft}
\end{equation}
where $\odot$ denotes elementwise multiplication. Restricting to $e_M$ preserves the MAE principle of supervising only what was masked.

\paragraph{Fourier amplitude loss.}
We penalize the magnitude of spectral error:
\begin{equation}
\mathcal{L}_{\text{FFT}}
=
\frac{1}{HW}
\sum_{\omega_u,\omega_v}
\big|E_M(\omega_u,\omega_v)\big|.
\label{eq:fft_loss}
\end{equation}
This term discourages spectral artifacts and encourages recovery of the missing high-frequency content that is important for sharp edges.

\paragraph{Radially-averaged PSD loss.}
Let $S_M(\omega_u,\omega_v)=|E_M(\omega_u,\omega_v)|^2$ be the power spectrum of the masked error.
To compare spectral structure at different spatial scales, we compute a radially-averaged PSD by binning frequencies into rings
$\{\mathcal{B}_r\}_{r=1}^{R}$:
\begin{equation}
\text{PSD}_M(r)
=
\frac{1}{|\mathcal{B}_r|}
\sum_{(\omega_u,\omega_v)\in\mathcal{B}_r} S_M(\omega_u,\omega_v).
\label{eq:psd_def}
\end{equation}
We then penalize the discrepancy between the reconstruction and ground truth PSDs on the masked region:
\begin{equation}
\mathcal{L}_{\text{PSD}}
=
\frac{1}{R}
\sum_{r=1}^{R}
\left|
\text{PSD}(\hat{x}_M)(r)
-
\text{PSD}(x_M)(r)
\right|,
\qquad
x_M=M\odot x,\;\; \hat{x}_M=M\odot \hat{x}.
\label{eq:psd_loss}
\end{equation}

\paragraph{Stage~3 objective.}
We define the frequency refinement loss and the final pretraining objective as:
\begin{align}
\mathcal{L}_{\text{freq}}
&=
\mathcal{L}_{\text{FFT}} + \eta\,\mathcal{L}_{\text{PSD}},
\label{eq:freq_loss}\\
\mathcal{L}_{\text{final}}
&=
\mathcal{L}_{\text{joint}} + \nu\,\mathcal{L}_{\text{freq}},
\label{eq:final_obj}
\end{align}
with weights $\eta$ and $\nu$ controlling the PSD contribution and the overall influence of frequency refinement.

\paragraph{Outcome of pretraining.}
Across the three stages, the model learns: (i) strong spatial priors for SEM imagery via masked reconstruction,
(ii) content-adaptive specialization via expert routing and distillation, and (iii) physically meaningful spectral consistency
that better preserves SEM-relevant frequency cues.
The resulting $\sim$1.7B-parameter SEM foundation model provides a robust backbone that can be efficiently adapted to multiple
downstream SEM imaging and measurement tasks.

\section{Downstream Task: Defocus-to-Focus Image Translation (Fig~\ref{fig:sem_fm_dt})}

After foundation model pretraining, we adapt our model to a downstream task of practical importance in SEM imaging: restoring in-focus images from defocused observations. Defocus artifacts commonly arise in automated microscopy due to sample drift, charging effects, or imperfect acquisition settings, and correcting them is essential for reliable downstream analysis.

Let \( x_d \in \mathbb{R}^{H \times W} \) denote a defocused SEM image and
\( x_f \in \mathbb{R}^{H \times W} \) its corresponding in-focus counterpart.
Our objective is to learn a mapping
\[
\hat{x}_f = g_\phi\!\left(f_\theta(x_d)\right),
\]
where \( f_\theta \) is the pretrained encoder (kept frozen) and
\( g_\phi \) is the MAE--MoE decoder fine-tuned for restoration.

To evaluate the sample efficiency enabled by the foundation model, we fine-tune the decoder using an extremely small paired dataset
\(
\mathcal{D}_{\text{train}} = \{(x_d^{(i)}, x_f^{(i)})\}_{i=1}^{M}
\),
with \( M = 2 \) image pairs.
Despite this severe data scarcity, the pretrained encoder provides sufficiently rich representations to enable stable adaptation.

\subsection{Training Objective}
\label{sec:training_objective}

We fine-tune the restoration model using a weighted combination of three complementary reconstruction terms that jointly promote (i) pixel-level fidelity, (ii) accurate recovery of sharp boundaries, and (iii) suppression of spurious high-frequency artifacts. The overall objective is
\begin{equation}
\mathcal{L}_{\text{total}}
=
\mathcal{L}_{\text{Charb}}
+
\lambda_{\text{e}}\,\mathcal{L}_{\text{edge}}
+
\lambda_{\text{tv}}\,\mathcal{L}_{\text{TV}},
\label{eq:total_loss}
\end{equation}
where $\lambda_{\text{e}}$ and $\lambda_{\text{tv}}$ control the relative emphasis on edge preservation and smoothness regularization, respectively.

\paragraph{Charbonnier Reconstruction Loss. \cite{charbonnier1994two}}
To enforce robust pixel-wise agreement between the predicted focused image $\hat{x}_f$ and the ground-truth focused image $x_f$, we use the Charbonnier loss (a differentiable approximation to $\ell_1$):
\begin{equation}
\mathcal{L}_{\text{Charb}}
=
\sum_{u,v}
\sqrt{ \left( \hat{x}_f(u,v) - x_f(u,v) \right)^2 + \epsilon^2 },
\label{eq:charb}
\end{equation}
where $\epsilon>0$ is a small constant for numerical stability. Compared to $\ell_2$, the Charbonnier penalty reduces sensitivity to outliers and rare intensity spikes, and compared to a non-smooth $\ell_1$, it yields stable gradients near zero. This is especially beneficial for SEM restoration, where signal-dependent noise and local contrast fluctuations can otherwise dominate optimization.

\paragraph{Edge (Gradient) Loss. \cite{mathieu2015deep}}
Pixel losses alone can lead to overly smooth reconstructions that attenuate fine boundaries. To explicitly preserve structural transitions (e.g., resist edges and line boundaries), we include a gradient consistency term:
\begin{equation}
\mathcal{L}_{\text{edge}}
=
\left\| \nabla_x \hat{x}_f - \nabla_x x_f \right\|_1
+
\left\| \nabla_y \hat{x}_f - \nabla_y x_f \right\|_1,
\label{eq:edge_loss}
\end{equation}
where $\nabla_x$ and $\nabla_y$ denote horizontal and vertical finite-difference operators. This loss aligns the edge responses of $\hat{x}_f$ with those of $x_f$, encouraging accurate recovery of sharp features and reducing boundary blurring that is detrimental to downstream SEM metrology.

\paragraph{Total Variation Regularization. \cite{rudin1992nonlinear}}
Finally, we regularize the prediction with anisotropic total variation (TV) to discourage local oscillations and isolated artifacts while retaining major edges:
\begin{equation}
\mathcal{L}_{\text{TV}}
=
\sum_{u,v}
\left(
\left| \hat{x}_f(u+1,v) - \hat{x}_f(u,v) \right|
+
\left| \hat{x}_f(u,v+1) - \hat{x}_f(u,v) \right|
\right).
\label{eq:tv}
\end{equation}
TV promotes piecewise-smooth solutions and helps suppress residual high-frequency noise that may persist after deblurring. In our extremely low-data adaptation setting, this regularizer also stabilizes fine-tuning by preventing the model from fitting noise patterns, while remaining complementary to the edge loss in Eq.~\ref{eq:edge_loss}, which preserves legitimate sharp transitions.

\subsection{Training Setup and Discussion}

During fine-tuning, the encoder \( f_\theta \) remains completely frozen and only the decoder \( g_\phi \) is updated.
This lightweight adaptation strategy exploits the semantic and structural priors learned during large-scale self-supervised pretraining, enabling rapid convergence from as few as one or two paired examples.

The chosen loss functions are intentionally simple and interpretable.
The Charbonnier loss enforces global reconstruction fidelity,
the edge loss preserves sharp structural transitions,
and total variation regularization suppresses noise while maintaining piecewise smoothness.
Together, these losses provide a strong and stable optimization objective without introducing task-specific complexity.

Importantly, our goal is not to engineer a specialized restoration network, but to demonstrate the effectiveness and generalizability of the pretrained foundation model.
More sophisticated objectives—such as adversarial losses, diffusion-based refinement, or physics-informed priors—could be incorporated on top of our framework.
However, even with this minimal loss design, the model achieves high-quality defocus-to-focus restoration with only a few minutes of fine-tuning, highlighting the strength and adaptability of the learned representations.

\paragraph{Synthetic defocus and noise modeling.}
Paired real defocus–focus SEM images are extremely scarce in practice, as acquiring multiple precisely aligned focus settings for the same field of view is time-consuming, instrument-dependent, and often infeasible during routine SEM operation \cite{batten2000autofocusing, lee2021robust}. To enable scalable training under this constraint, we generate synthetic defocus–focus pairs using a physics-inspired forward image formation model.
Given a focused image $x(\mathbf{r})$, the corresponding defocused observation $y(\mathbf{r})$ is modeled as
\begin{equation}
\label{fr-model}
y(\mathbf{r}) = \mathcal{N}\!\left( a \cdot \big(x * h_{\boldsymbol{\theta}}\big)(\mathbf{r}) + b \right)
\end{equation}

where $*$ denotes convolution, $h_{\boldsymbol{\theta}}$ is the SEM point spread function (PSF) parameterized by defocus and astigmatism parameters $\boldsymbol{\theta}$, $(a,b)$ model global intensity scaling and offset, and $\mathcal{N}(\cdot)$ denotes signal-dependent noise.
The ideal diffraction-limited PSF is modeled using an Airy pattern \cite{goodman1969introduction, wolf2007optics},
\[
h_{\text{Airy}}(r) = \left[ \frac{2 J_1(\pi r)}{\pi r} \right]^{\beta},
\]
where $J_1(\cdot)$ is the first-order Bessel function of the first kind and $r$ is the radial distance from the optical axis. To capture practical SEM aberrations, we generalize this model to an anisotropic and rotated PSF by defining
\[
r = \sqrt{\left(\frac{x'}{R_x}\right)^2 + \left(\frac{y'}{R_y}\right)^2},
\quad
\begin{pmatrix}
x' \\ y'
\end{pmatrix}
=
\begin{pmatrix}
\cos\theta & \sin\theta \\
-\sin\theta & \cos\theta
\end{pmatrix}
\begin{pmatrix}
x \\ y
\end{pmatrix},
\]
where $R_x$ and $R_y$ control elliptical defocus along orthogonal axes (modeling astigmatism), and $\theta$ specifies the astigmatism orientation \cite{shechtman2014optimal}. Following prior SEM modeling practice, we additionally raise the Airy response to a power $\beta$ to flexibly capture deviations from the ideal diffraction-limited response.
Noise is modeled as a combination of Poisson and Gaussian components \cite{chong2023m, mevenkamp2015poisson, mannam2022real},
\[
\mathcal{N}(z) = \frac{1}{\text{dose}}\,\text{Poisson}(z \cdot \text{dose}) + \mathcal{N}(0, \sigma^2),
\]
where the Poisson term models electron counting statistics governed by the effective dose, and the Gaussian term with variance $\sigma^2$ accounts for additive electronic readout noise.
All PSF and noise parameters $(R_x, R_y, \beta, \theta, a, b, \sigma, \text{dose})$ are sampled using the same ranges as those employed for the baseline methods, ensuring a consistent and fair synthetic data generation protocol across all models. By randomizing these parameters during training, we expose the network to a broad yet physically plausible range of defocus and noise conditions, enabling robust generalization to real defocused SEM images despite the absence of paired real training data.

\section{Conclusion}

We introduced the first large-scale SEM foundation model and demonstrated its effectiveness for defocus-to-focus image restoration under realistic experimental conditions. By combining large-scale self-supervised pretraining on diverse SEM data with lightweight domain adaptation and physics-aware degradation modeling, our method bridges the gap between synthetic training data and real-world SEM deployment. Extensive comparisons against classical deconvolution, denoising-based approaches, and specialized deep learning models show consistent improvements across perceptual quality metrics (PSNR, SSIM, LPIPS, NIQE) as well as domain-critical metrology measures. Crucially, these improvements are achieved without requiring extensive real focused training data, making the approach practical for routine SEM workflows. Beyond defocus correction, this work establishes a foundation for a broader class of SEM-aware learning tasks, including denoising, super-resolution, segmentation, and measurement-aware reconstruction. We believe that SEM-specific foundation models represent a scalable and unifying paradigm for scientific imaging, enabling robust, data-efficient, and physically grounded solutions across microscopy and related domains.

\backmatter

\bmhead{Supplementary information}

We report the real data acquistion details corresponding to Table~\ref{tab:real_one_pair}.

\paragraph{Fabrication of block copolymer (BCP)-derived inorganic nanostructures}

The BCP-derived inorganic nanostructures used in this study were manufactured following previously reported procedures~\cite{lee2024effects} that include block copolymer self-assembly, vapor-phase infiltration, and subsequent polymer removal and consolidation steps. Self-assembled BCP patterns were prepared using a polystyrene-block-poly(methyl methacrylate) (PS-\emph{b}-PMMA) system. PS-\emph{b}-PMMA ($M_n = 105$~kg~mol$^{-1}$, block ratio 47:58) was purchased from Polymer Source. Silicon substrates were first cleaned by oxygen plasma treatment (20~W, 100~mTorr, 1~min) using a reactive ion etcher (March CS-1701). To establish neutral wetting conditions, a random copolymer brush layer of polystyrene-\emph{random}-poly(methyl methacrylate) (PS-\emph{r}-PMMA, $M_n = 9.2$~kg~mol$^{-1}$, block ratio 61:39, provided by The Dow Chemical Company) was spin-coated from a 1~wt\% solution in propylene glycol methyl ether acetate (PGMEA). The brush layer was thermally annealed on a nitrogen-blanketed hot plate at 250~$^\circ$C for 5~min, followed by rinsing with toluene to remove ungrafted chains. A lamellar-phase PS-\emph{b}-PMMA thin film (1~wt\% in toluene) was subsequently spin-coated onto the brush-treated substrate and annealed under a nitrogen atmosphere at 250~$^\circ$C for 5~min to induce microphase separation and formation of vertically oriented lamellar nanostructures.

Subsequently, vapor-phase infiltration was performed in a commercial atomic layer deposition (ALD) system (Veeco, Savannah S-200) operated at 85 °C using trimethylaluminum (TMA) and diethylzinc (DEZ) as metal–organic precursors, with water serving as the oxidant. During each precursor exposure, the ALD chamber was operated under static vacuum conditions, in which the chamber was isolated from the pump during precursor dosing and held under static conditions for the prescribed exposure time to promote efficient precursor diffusion and infiltration into the polymer domains. An initial AlOx priming step was performed using a single TMA exposure followed by water exposure to facilitate subsequent metal infiltration. Following the priming step, ZnOx infiltration was carried out using a microdose protocol, in which DEZ was introduced through multiple short pulses distributed over an extended static exposure period, followed by water exposure under similar conditions. This sequence was repeated for a total of six infiltration

cycles. After each exposure step, the chamber was purged with nitrogen under dynamic pumping conditions to remove excess precursor and byproducts, completing each infiltration cycle. Following infiltration, the organic polymer matrix was removed by oxygen plasma etching (20 W, 100 mTorr, 5 min, room temperature). The resulting inorganic framework was further consolidated, and residual carbon impurities were removed by oxygen rapid thermal processing (RTP) at 600 °C for 5 min using an RTP system (Modular Process Technology, RTP-600S), yielding well-defined BCP-derived metal oxide nanostructures.

SEM imaging

SEM imaging was performed using a field-emission SEM (Hitachi S-4800). SEM images were acquired at multiple magnifications sufficient to clearly resolve the BCP-derived nanostructures. For each magnification, images were systematically collected under varying focus conditions (under-focus, in-focus, and over-focus) and scan speeds (fast, medium, and slow). All SEM images were collected with a resolution of 1280 × 960 pixels to ensure sufficient image quality for machine learning model training and evaluation.

\bmhead{Acknowledgements}

This research is supported by the U.S. Department of Energy Office of Science Accelerate Initiative Award 2023-BNL-NC033-Fund. The authors thank Dr. Dario Goldfarb at IBM Research for providing the EUV pattern samples.





\bibliography{sn-bibliography}

@inproceedings{he2022masked,
  title={Masked autoencoders are scalable vision learners},
  author={He, Kaiming and Chen, Xinlei and Xie, Saining and Li, Yanghao and Doll{\'a}r, Piotr and Girshick, Ross},
  booktitle={Proceedings of the IEEE/CVF conference on computer vision and pattern recognition},
  pages={16000--16009},
  year={2022}
}

@incollection{postek2001critical,
  title={Critical-dimension metrology and the scanning electron microscope},
  author={Postek, Michael T and Vlad{\'a}r, Andr{\'a}s E},
  booktitle={Handbook of Silicon Semiconductor Metrology},
  pages={244--275},
  year={2001},
  publisher={CRC Press}
}

@article{orji2018metrology,
  title={Metrology for the next generation of semiconductor devices},
  author={Orji, Ndubuisi G and Badaroglu, Mustafa and Barnes, Bryan M and Beitia, Carlos and Bunday, Benjamin D and Celano, Umberto and Kline, Regis J and Neisser, Mark and Obeng, Yaw and Vladar, AE},
  journal={Nature electronics},
  volume={1},
  number={10},
  pages={532--547},
  year={2018},
  publisher={Nature Publishing Group UK London}
}

@inproceedings{kumar2025resist,
  title={Resist level critical dimension SEM metrology: challenges and developments for EUV photomasks},
  author={Kumar, Deepan Kishore and Seeger, Adam A and Ganti, Prathyusha and Mohan, Varun and Straney, Patrick and Rice, Zachary and Sundaramurthy, Arvind and Tavassoli, Malahat},
  booktitle={Photomask Technology 2025},
  volume={13687},
  pages={34--44},
  year={2025},
  organization={SPIE}
}

@inproceedings{lorusso2022metrology,
  title={Metrology of thin resist for high NA EUVL},
  author={Lorusso, Gian Francesco and Beral, Christophe and Bogdanowicz, Janusz and De Simone, Danilo and Hasan, Mahmudul and Jehoul, Christiane and Moussa, Alain and Saib, Mohamed and Zidan, Mohamed and Severi, Joren and others},
  booktitle={Metrology, Inspection, and Process Control XXXVI},
  volume={12053},
  pages={229--240},
  year={2022},
  organization={SPIE}
}

@article{lorusso2018unbiased,
  title={Unbiased roughness measurements: subtracting out SEM effects},
  author={Lorusso, Gian F and Rutigliani, Vito and Van Roey, Frieda and Mack, Chris A},
  journal={Microelectronic Engineering},
  volume={190},
  pages={33--37},
  year={2018},
  publisher={Elsevier}
}

@inproceedings{orji2021spectral,
  title={Spectral analysis of line edge and line width roughness using wavelets},
  author={Orji, Ndubuisi G},
  booktitle={Metrology, Inspection, and Process Control for Semiconductor Manufacturing XXXV},
  volume={11611},
  pages={255--266},
  year={2021},
  organization={SPIE}
}

@article{schubert2024deepfocus,
  title={DeepFocus: Fast focus and astigmatism correction for electron microscopy},
  author={Schubert, Philipp Johannes and Saxena, Rangoli and Kornfeld, Joergen},
  journal={Nature Communications},
  volume={15},
  number={1},
  pages={948},
  year={2024},
  publisher={Nature Publishing Group UK London}
}

@article{maraghechi2019correction,
  title={Correction of scanning electron microscope imaging artifacts in a novel digital image correlation framework},
  author={Maraghechi, S and Hoefnagels, JPM and Peerlings, RHJ and Roko{\v{s}}, O and Geers, MGD},
  journal={Experimental mechanics},
  volume={59},
  number={4},
  pages={489--516},
  year={2019},
  publisher={Springer}
}

@inproceedings{abaidi2025analytical,
  title={Analytical methods for SEM image enhancement: noise and charging effect reduction for precise contour extraction},
  author={Abaidi, Mohamed and Yang, XiaoChun and Fang, Hawren and Clifford, Chris and Meng, Renyang and Gillijns, Werner},
  booktitle={40th European Mask and Lithography Conference (EMLC 2025)},
  volume={13787},
  pages={255--270},
  year={2025},
  organization={SPIE}
}

@inproceedings{chung2025true,
  title={True metrology with reduced resist shrinkage effect for process window optimization and EUV stochastic effects analysis},
  author={Chung, No-Young and Harari, Yoav},
  booktitle={Metrology, Inspection, and Process Control XXXIX},
  volume={13426},
  pages={134260C},
  year={2025},
  organization={SPIE}
}

@article{park2025deep,
  title={Deep learning denoising enables rapid SEM imaging under charging conditions for FE SEM, CD SEM, and review SEM},
  author={Park, Hyungjoo and Oh, Beom-Seok and Jang, Kuk Jin},
  journal={Scientific Reports},
  year={2025},
  publisher={Nature Publishing Group UK London}
}

@misc{MyScopeSEMArtefacts,
  title  = {Image artefacts and trouble-shooting -- SEM},
  author = {{MyScope Training}},
  year   = {n.d.},
  note   = {Educational resource on common SEM artifacts including charging},
  url    = {https://myscope.training/SEM_Image_artefacts_and_trouble_shooting}
}

@article{dosovitskiy2020image,
  title={An image is worth 16x16 words: Transformers for image recognition at scale},
  author={Dosovitskiy, Alexey and Beyer, Lucas and Kolesnikov, Alexander and Weissenborn, Dirk and Zhai, Xiaohua and Unterthiner, Thomas and Dehghani, Mostafa and Minderer, Matthias and Heigold, Georg and Gelly, Sylvain and others},
  journal={arXiv preprint arXiv:2010.11929},
  year={2020}
}

@article{jaiswal2020survey,
  title={A survey on contrastive self-supervised learning},
  author={Jaiswal, Ashish and Babu, Ashwin Ramesh and Zadeh, Mohammad Zaki and Banerjee, Debapriya and Makedon, Fillia},
  journal={Technologies},
  volume={9},
  number={1},
  pages={2},
  year={2020},
  publisher={MDPI}
}

@article{bommasani2021opportunities,
  title={On the opportunities and risks of foundation models},
  author={Bommasani, Rishi and Hudson, Drew A and Adeli, Ehsan and Altman, Russ and Arora, Simran and von Arx, Sydney and Bernstein, Michael S and Bohg, Jeannette and Bosselut, Antoine and Brunskill, Emma and others},
  journal={arXiv preprint arXiv:2108.07258},
  year={2021}
}

@inproceedings{caron2021emerging,
  title={Emerging properties in self-supervised vision transformers},
  author={Caron, Mathilde and Touvron, Hugo and Misra, Ishan and J{\'e}gou, Herv{\'e} and Mairal, Julien and Bojanowski, Piotr and Joulin, Armand},
  booktitle={Proceedings of the IEEE/CVF international conference on computer vision},
  pages={9650--9660},
  year={2021}
}

@inproceedings{chen2021empirical,
  title={An empirical study of training self-supervised vision transformers},
  author={Chen, Xinlei and Xie, Saining and He, Kaiming},
  booktitle={Proceedings of the IEEE/CVF international conference on computer vision},
  pages={9640--9649},
  year={2021}
}

@article{shazeer2017outrageously,
  title={Outrageously large neural networks: The sparsely-gated mixture-of-experts layer},
  author={Shazeer, Noam and Mirhoseini, Azalia and Maziarz, Krzysztof and Davis, Andy and Le, Quoc and Hinton, Geoffrey and Dean, Jeff},
  journal={arXiv preprint arXiv:1701.06538},
  year={2017}
}

@article{fedus2022switch,
  title={Switch transformers: Scaling to trillion parameter models with simple and efficient sparsity},
  author={Fedus, William and Zoph, Barret and Shazeer, Noam},
  journal={Journal of Machine Learning Research},
  volume={23},
  number={120},
  pages={1--39},
  year={2022}
}

@article{loshchilov2017decoupled,
  title={Decoupled weight decay regularization},
  author={Loshchilov, Ilya and Hutter, Frank},
  journal={arXiv preprint arXiv:1711.05101},
  year={2017}
}

@article{wang2004image,
  title={Image quality assessment: from error visibility to structural similarity},
  author={Wang, Zhou and Bovik, Alan C and Sheikh, Hamid R and Simoncelli, Eero P},
  journal={IEEE transactions on image processing},
  volume={13},
  number={4},
  pages={600--612},
  year={2004},
  publisher={IEEE}
}

@article{sara2019image,
  title={Image quality assessment through FSIM, SSIM, MSE and PSNR—a comparative study},
  author={Sara, Umme and Akter, Morium and Uddin, Mohammad Shorif and others},
  journal={Journal of Computer and Communications},
  volume={7},
  number={3},
  pages={8--18},
  year={2019}
}

@article{mittal2012making,
  title={Making a “completely blind” image quality analyzer},
  author={Mittal, Anish and Soundararajan, Rajiv and Bovik, Alan C},
  journal={IEEE Signal processing letters},
  volume={20},
  number={3},
  pages={209--212},
  year={2012},
  publisher={IEEE}
}

@inproceedings{zhang2018unreasonable,
  title={The unreasonable effectiveness of deep features as a perceptual metric},
  author={Zhang, Richard and Isola, Phillip and Efros, Alexei A and Shechtman, Eli and Wang, Oliver},
  booktitle={Proceedings of the IEEE conference on computer vision and pattern recognition},
  pages={586--595},
  year={2018}
}

@article{azarnouche2012unbiased,
  title={Unbiased line width roughness measurements with critical dimension scanning electron microscopy and critical dimension atomic force microscopy},
  author={Azarnouche, L and Pargon, E and Menguelti, K and Fouchier, M and Fuard, D and Gouraud, P and Verove, C and Joubert, O},
  journal={Journal of Applied Physics},
  volume={111},
  number={8},
  year={2012},
  publisher={AIP Publishing}
}

@article{richardson1972bayesian,
  title={Bayesian-based iterative method of image restoration},
  author={Richardson, William Hadley},
  journal={Journal of the optical society of America},
  volume={62},
  number={1},
  pages={55--59},
  year={1972},
  publisher={Optical Society of America}
}

@article{lucy1974iterative,
  title={An iterative technique for the rectification of observed distributions},
  author={Lucy, Leon B},
  journal={Astronomical Journal, Vol. 79, p. 745 (1974)},
  volume={79},
  pages={745},
  year={1974}
}

@book{wiener1949extrapolation,
  title={Extrapolation, interpolation, and smoothing of stationary time series: with engineering applications},
  author={Wiener, Norbert},
  year={1949},
  publisher={The MIT press}
}

@article{dabov2007image,
  title={Image denoising by sparse 3-D transform-domain collaborative filtering},
  author={Dabov, Kostadin and Foi, Alessandro and Katkovnik, Vladimir and Egiazarian, Karen},
  journal={IEEE Transactions on image processing},
  volume={16},
  number={8},
  pages={2080--2095},
  year={2007},
  publisher={IEEE}
}

@article{lehtinen2018noise2noise,
  title={Noise2Noise: Learning image restoration without clean data},
  author={Lehtinen, Jaakko and Munkberg, Jacob and Hasselgren, Jon and Laine, Samuli and Karras, Tero and Aittala, Miika and Aila, Timo},
  journal={arXiv preprint arXiv:1803.04189},
  year={2018}
}

@inproceedings{krull2019noise2void,
  title={Noise2void-learning denoising from single noisy images},
  author={Krull, Alexander and Buchholz, Tim-Oliver and Jug, Florian},
  booktitle={Proceedings of the IEEE/CVF conference on computer vision and pattern recognition},
  pages={2129--2137},
  year={2019}
}

@article{na2021deep,
  title={Deep learning-based discriminative refocusing of scanning electron microscopy images for materials science},
  author={Na, Juwon and Kim, Gyuwon and Kang, Seong-Hoon and Kim, Se-Jong and Lee, Seungchul},
  journal={Acta Materialia},
  volume={214},
  pages={116987},
  year={2021},
  publisher={Elsevier}
}

@article{russakovsky2015imagenet,
  title={Imagenet large scale visual recognition challenge},
  author={Russakovsky, Olga and Deng, Jia and Su, Hao and Krause, Jonathan and Satheesh, Sanjeev and Ma, Sean and Huang, Zhiheng and Karpathy, Andrej and Khosla, Aditya and Bernstein, Michael and others},
  journal={International journal of computer vision},
  volume={115},
  number={3},
  pages={211--252},
  year={2015},
  publisher={Springer}
}

@article{rousseeuw1987silhouettes,
  title={Silhouettes: a graphical aid to the interpretation and validation of cluster analysis},
  author={Rousseeuw, Peter J},
  journal={Journal of computational and applied mathematics},
  volume={20},
  pages={53--65},
  year={1987},
  publisher={Elsevier}
}

@article{davies2009cluster,
  title={A cluster separation measure},
  author={Davies, David L and Bouldin, Donald W},
  journal={IEEE transactions on pattern analysis and machine intelligence},
  number={2},
  pages={224--227},
  year={2009},
  publisher={Ieee}
}

@article{calinski1974dendrite,
  title={A dendrite method for cluster analysis},
  author={Cali{\'n}ski, Tadeusz and Harabasz, Jerzy},
  journal={Communications in Statistics-theory and Methods},
  volume={3},
  number={1},
  pages={1--27},
  year={1974},
  publisher={Taylor \& Francis}
}

@inproceedings{charbonnier1994two,
  title={Two deterministic half-quadratic regularization algorithms for computed imaging},
  author={Charbonnier, Pierre and Blanc-Feraud, Laure and Aubert, Gilles and Barlaud, Michel},
  booktitle={Proceedings of 1st international conference on image processing},
  volume={2},
  pages={168--172},
  year={1994},
  organization={IEEE}
}

@article{mathieu2015deep,
  title={Deep multi-scale video prediction beyond mean square error},
  author={Mathieu, Michael and Couprie, Camille and LeCun, Yann},
  journal={arXiv preprint arXiv:1511.05440},
  year={2015}
}

@article{rudin1992nonlinear,
  title={Nonlinear total variation based noise removal algorithms},
  author={Rudin, Leonid I and Osher, Stanley and Fatemi, Emad},
  journal={Physica D: nonlinear phenomena},
  volume={60},
  number={1-4},
  pages={259--268},
  year={1992},
  publisher={Elsevier}
}

@inproceedings{mochi2020open,
  title={Open-source software for SEM metrology},
  author={Mochi, Iacopo and Vockenhuber, Michaela and Allenet, Timoth{\'e}e and Ekinci, Yasin},
  booktitle={Photomask Technology 2020},
  volume={11518},
  pages={58--67},
  year={2020},
  organization={SPIE}
}

@inproceedings{mochi2021contacts,
  title={Contacts and lines SEM image metrology with SMILE},
  author={Mochi, Iacopo and Vockenhuber, Michaela and Allenet, Timoth{\'e}e and Ekinci, Yasin},
  booktitle={Photomask Technology 2021},
  volume={11855},
  pages={1185502},
  year={2021},
  organization={SPIE}
}

@inproceedings{develioglu2023euv,
  title={The EUV lithography resist screening activities in H2-2022},
  author={Develioglu, Aysegul and Allenet, Tim P and Vockenhuber, Michaela and van Lent-Protasova, Lidia and Mochi, Iacopo and Ekinci, Yasin and Kazazis, Dimitrios},
  booktitle={Advances in Patterning Materials and Processes XL},
  volume={12498},
  pages={9--17},
  year={2023},
  organization={SPIE}
}

@article{batten2000autofocusing,
  title={Autofocusing and astigmatism correction in the scanning electron microscope},
  author={Batten, Christopher F},
  journal={Mphill thesis, University of Cambridge},
  volume={89},
  year={2000}
}

@article{lee2021robust,
  title={Robust autofocusing for scanning electron microscopy based on a dual deep learning network},
  author={Lee, Woojin and Nam, Hyeong Soo and Kim, Young Gon and Kim, Yong Ju and Lee, Jun Hee and Yoo, Hongki},
  journal={Scientific Reports},
  volume={11},
  number={1},
  pages={20933},
  year={2021},
  publisher={Nature Publishing Group UK London}
}

@misc{goodman1969introduction,
  title={Introduction to Fourier optics},
  author={Goodman, Joseph W and Cox, Mary E},
  year={1969},
  publisher={American Institute of Physics}
}

@article{wolf2007optics,
  title={The optics of microscope image formation},
  author={Wolf, David E},
  journal={Methods in cell biology},
  volume={81},
  pages={11--42},
  year={2007},
  publisher={Elsevier}
}

@article{shechtman2014optimal,
  title={Optimal point spread function design for 3D imaging},
  author={Shechtman, Yoav and Sahl, Steffen J and Backer, Adam S and Moerner, William E},
  journal={Physical review letters},
  volume={113},
  number={13},
  pages={133902},
  year={2014},
  publisher={APS}
}

@article{chong2023m,
  title={M-Denoiser: Unsupervised image denoising for real-world optical and electron microscopy data},
  author={Chong, Xiaoya and Cheng, Min and Fan, Wenqi and Li, Qing and Leung, Howard},
  journal={Computers in Biology and Medicine},
  volume={164},
  pages={107308},
  year={2023},
  publisher={Elsevier}
}

@article{mevenkamp2015poisson,
  title={Poisson noise removal from high-resolution STEM images based on periodic block matching},
  author={Mevenkamp, Niklas and Binev, Peter and Dahmen, Wolfgang and Voyles, Paul M and Yankovich, Andrew B and Berkels, Benjamin},
  journal={Advanced Structural and Chemical Imaging},
  volume={1},
  number={1},
  pages={3},
  year={2015},
  publisher={Springer}
}

@article{mannam2022real,
  title={Real-time image denoising of mixed Poisson--Gaussian noise in fluorescence microscopy images using ImageJ},
  author={Mannam, Varun and Zhang, Yide and Zhu, Yinhao and Nichols, Evan and Wang, Qingfei and Sundaresan, Vignesh and Zhang, Siyuan and Smith, Cody and Bohn, Paul W and Howard, Scott S},
  journal={Optica},
  volume={9},
  number={4},
  pages={335--345},
  year={2022},
  publisher={Optica Publishing Group}
}

@article{lepikhin2020gshard,
  title={Gshard: Scaling giant models with conditional computation and automatic sharding},
  author={Lepikhin, Dmitry and Lee, HyoukJoong and Xu, Yuanzhong and Chen, Dehao and Firat, Orhan and Huang, Yanping and Krikun, Maxim and Shazeer, Noam and Chen, Zhifeng},
  journal={arXiv preprint arXiv:2006.16668},
  year={2020}
}

@article{lee2024effects,
  title={Effects of alumina priming on the electrical properties of ZnO nanostructures derived from vapor-phase infiltration into self-assembled block copolymer thin films},
  author={Lee, Won-Il and Subramanian, Ashwanth and Kisslinger, Kim and Tiwale, Nikhil and Nam, Chang-Yong},
  journal={Materials Advances},
  volume={5},
  number={14},
  pages={5698--5708},
  year={2024},
  publisher={Royal Society of Chemistry}
}

\end{document}